\documentclass[11pt]{article}

\usepackage{bbm}
\usepackage{amsmath,amsthm,amssymb}
\usepackage{graphicx}
\usepackage{tikz}
\usepackage[utf8]{inputenc}
\usepackage{pgfplots}
\usepackage{hyperref}
\usepackage{setspace}
\usepackage{pict2e,color}
\usepackage{multirow}
\usepackage{booktabs}
\usepackage{amsfonts}
\usepackage{verbatim} 
\usepackage{sectsty}
\usepackage{url}
\usepackage{empheq}
\usepackage[normalem]{ulem}
\usepackage[titletoc,title]{appendix}
\usepackage[flushleft]{threeparttable}
\usepackage{soul} 
\usepackage{natbib}
\usepackage{algorithm,algorithmic}
\usepackage{microtype}
\usepackage{subfigure}
\usepackage{pbox}

\usepackage{etoolbox}
\preto\subequations{\ifhmode\unskip\fi}
\usepackage{booktabs}
\usepackage{multirow}

\theoremstyle{definition}

\newtheorem{theorem}{Theorem}
\newtheorem{lemma}{Lemma}
\newtheorem{proposition}{Proposition}

\newtheorem{remark}{Remark}

\usepackage{lastpage}
\allowdisplaybreaks

\title{On the Global Convergence of Risk-Averse Natural Policy Gradient Methods with Expected Conditional Risk Measures}
\author{Xian Yu\thanks{Corresponding author; Department of Integrated Systems Engineering, The Ohio State University, Columbus, OH, USA, Email: {\tt yu.3610@osu.edu};}~~~and Lei Ying\thanks{Department of Electrical Engineering and Computer Science, University of Michigan, Ann Arbor, MI, USA, Email: {\tt leiying@umich.edu}.}}
\date{}

\begin{document}

\maketitle

\begin{abstract}
Risk-sensitive reinforcement learning (RL) has become a popular tool for controlling the risk of uncertain outcomes and ensuring reliable performance in highly stochastic sequential decision-making problems. While it has been shown that policy gradient methods can find globally optimal policies in the risk-neutral setting \citep{mei2020global,agarwal2021theory,cen2022fast,bhandari2024global}, it remains unclear if the risk-averse variants enjoy the same global convergence guarantees. In this paper, we consider a class of dynamic time-consistent risk measures, named Expected Conditional Risk Measures (ECRMs), and derive natural policy gradient (NPG) updates for ECRMs-based RL problems. {\color{black}We provide global optimality and iteration complexity of the proposed risk-averse NPG algorithm with softmax parameterization and entropy regularization under both exact and inexact policy evaluation. Furthermore, we test our risk-averse NPG algorithm on a stochastic Cliffwalk environment to demonstrate the efficacy of our method.}
\end{abstract}




~\\
{\bf Keywords:} Reinforcement Learning, Coherent Risk Measures, Natural Policy Gradient, Global Convergence

\section{Introduction}
\label{sec:intro}
Sequential decision-making problems appear ubiquitously in real-world applications across different fields, where a decision-maker interacts with a stochastic environment and collects reward/cost over time. This type of problem can often be modeled as Markov Decision Processes (MDPs) and has been extensively studied in the reinforcement learning (RL) literature \cite{puterman2014markov,sutton2018reinforcement}.
In risk-neutral RL, the decision-maker seeks a policy that minimizes the expected total cost (or maximizes the expected total reward). However, minimizing the expected cost does not necessarily avoid the rare occurrences of undesirably high costs, and  
in high-stakes applications, we aim to evaluate and control the \textit{risk}. 
    The risk can be measured on the total cumulative cost or in a nested way, leading to static or dynamic risk measures, respectively. 
    {\color{black}While static risk measures are more intuitive, it has been shown that their globally optimal policies are generally history-dependent \citep{bauerle2011markov}. Because of this,}
    we consider a class of dynamic time-consistent risk measures, named expected conditional risk measures (ECRMs) \citep{homem2016risk}. Using a convex combination of expectation and Conditional-Value-at-Risk (CVaR) as the one-step conditional risk measure, \cite{yu2022risk} showed that the resulting ECRM is time-consistent and possesses a decomposable structure that allows us to reformulate the risk-sensitive RL problem as a risk-neutral counterpart. They further proved that the corresponding risk-averse Bellman operator is a contraction mapping, which guarantees the global convergence of value-based RL algorithms. In an earlier version of our work \citep{yu2023global}, we have shown that risk-averse policy gradient methods also possess global convergence guarantees for ECRM-based RL problems under direct and softmax parameterizations.
The major contributions of this paper are threefold. First, we apply ECRMs on infinite-horizon MDPs and propose risk-averse {\color{black}natural policy gradient (NPG) updates for ECRMs-based RL. Second, analogous to the risk-neutral case, we establish global optimality and iteration complexity for the proposed risk-averse NPG methods with softmax parameterization and entropy regularization under exact policy evaluation. Third, we study approximate NPG algorithms under inexact policy evaluation and analyze their convergence properties. These convergence results closely match the risk-neutral ones in \cite{cen2022fast}.}

\section{Preliminaries}
\label{sec:problem}
We consider an infinite horizon discounted MDP denoted by a tuple $M=(\mathcal{S},\mathcal{A},C, P,\gamma,\rho)$, where $\mathcal{S}$ is a finite state space, $\mathcal{A}$ is a finite action space, $C(s,a)\in[0,1]$ is a bounded and deterministic cost given state $s\in\mathcal{S}$ and action $a\in\mathcal{A}$, $P(\cdot|s,a)$ is a transition probability distribution, $\gamma\in(0,1)$ is a discount factor, and $\rho$ is an initial state distribution over $\mathcal{S}$.

A stationary Markov policy $\pi^{\theta}: \mathcal{S}\to\Delta(\mathcal{A})$ parameterized by $\theta$ specifies a probability distribution over the action space given each state $s\in\mathcal{S}$, where $\Delta(\cdot)$ denotes the probability simplex, i.e., $0\le\pi^{\theta}(a|s)\le 1,\ \sum_{a\in\mathcal{A}}\pi^{\theta}(a|s)=1,\ \forall s\in\mathcal{S},\ a\in\mathcal{A}$. 
A policy induces a distribution over trajectories $\{(s_t,a_t,C(s_t,a_t))\}_{t=1}^{\infty}$, where $s_1$ is drawn from the initial state distribution $\rho$, and for all time steps $t$, $a_t\sim \pi^{\theta}(\cdot|s_t),\ s_{t+1}\sim P(\cdot|s_t,a_t)$. The value function $V^{\pi^{\theta}}:\mathcal{S}\to\mathbb{R}$ is defined as the expectation of the total discounted cost starting at state $s$ and executing $\pi$, i.e.,
$V^{\pi^{\theta}}(s)=\mathbb{E}[\sum_{t=1}^{\infty}\gamma^{t-1}C(s_t,a_t)|\pi^{\theta}, s_1=s].$
We overload the notation and define $V^{\pi^{\theta}}(\rho)$ as the expected value under initial state distribution $\rho$, i.e., $V^{\pi^{\theta}}(\rho)=\mathbb{E}_{s_1\sim\rho}[V^{\pi_\theta}(s_1)]$.
The action-value (or Q-value) function $Q^{\pi^{\theta}}:\mathcal{S}\times\mathcal{A}\to\mathbb{R}$ is defined as $
    Q^{\pi^{\theta}}(s,a)=\mathbb{E}[\sum_{t=1}^{\infty}\gamma^{t-1}C(s_t,a_t)|\pi^{\theta}, s_1=s,a_1=a]$.

In risk-neutral RL, the goal is to find a policy $\pi^{\theta}$ that minimizes the expected total cost from the initial state distribution, i.e., 
$\min_{\theta\in\Theta}V^{\pi^{\theta}}(\rho)$ where $\{\pi^{\theta}|\theta\in\Theta\}$ is some class of parametric stochastic policies.
The famous theorem of \citet{bellman1959functional} shows that there exists a policy $\pi^*$ that simultaneously minimizes $V^{\pi^{\theta}}(s_1)$ for all states $s_1\in\mathcal{S}$. It is worth noting that $V^{\pi^{\theta}}(s)$ is non-convex in $\theta$, so the standard tools from convex optimization literature are not applicable. We refer interested readers to \citet{agarwal2021theory} for a non-convex example in Figure 1.

\paragraph{Notation.} Throughout the paper, we rewrite $C(s_t,a_t)$ as $c_t$ for all $t\ge 1$ and denote any vector $(a_1,\ldots,a_t)$ as $a_{[1,t]}$. 
Let $\mathbb{E}_{s_{t}}^{s_{t-1}}=\mathbb{E}_{s_{t}}[\cdot|{s_{t-1}}]$ denote the conditional expectation over $s_t$ conditioned on $s_{t-1}$. 

\subsection{Policy Gradient Methods}
PG algorithms have received lots of attention in the RL community due to their simple structure. The basic idea is to adjust the parameter $\theta$ of the policy in the gradient descent direction. 
The fundamental result underlying PG algorithms is the \textit{PG theorem} \citep{williams1992simple, sutton1999policy}, i.e., $\nabla_{\theta}V^{\pi^{\theta}}(s_1)=\frac{1}{1-\gamma}\mathbb{E}_{s\sim d_{s_1}^{\pi^{\theta}}}\mathbb{E}_{a\sim \pi^{\theta}(\cdot|s)}[\nabla_{\theta}\log\pi^{\theta}(a|s)Q^{\pi^{\theta}}(s,a)]$,
where the gradient is surprisingly simple and does not depend on the gradient of the state distribution.

Recently, \citet{mei2020global,agarwal2021theory,cen2022fast,bhandari2024global} demonstrate the global convergence of PG methods in a risk-neutral setting. This paper aims to extend the results to risk-averse objective functions with a class of dynamic time-consistent risk measures. Next, we first introduce the coherent one-step conditional risk measure used in this paper. 

\subsection{Coherent One-Step Conditional Risk Measures}
\label{sec:def-coherent}
Consider a probability space $(\Xi,\mathcal{F},P)$, and let $\mathcal{F}_1\subset\mathcal{F}_2\subset\ldots$ be sub-sigma-algebras of $\mathcal{F}$ such that each $\mathcal{F}_t$ corresponds to the information available up to (and including) stage $t$, with $\{Z_t\}_{t=1}^{\infty}$ being an adapted sequence of random variables. In this paper, we interpret random variables $Z_t$ as costs and the smaller the better. We assume that $\mathcal{F}_1=\{\emptyset,\Xi\}$ is the trivial sigma-algebra, and $Z_1$ is deterministic. Let $\mathcal{Z}_t$ denote the space of $\mathcal{F}_t$-measurable functions mapping from $\Xi$ to $\mathbb{R}$. 

For our problem, we consider a special class of coherent one-step conditional risk measures $\varrho_t^{s_{[1,t-1]}}$ mapping from $\mathcal{Z}_t$ to $\mathcal{Z}_{t-1}$, which is a convex combination of conditional expectation and Conditional Value-at-Risk (CVaR):
\begin{equation}
\varrho_t^{s_{[1,t-1]}}(c_t)=(1-\lambda)\mathbb{E}[c_t|s_{[1,t-1]}]+\lambda \text{CVaR}_{\alpha}[c_t|s_{[1,t-1]}],
\label{eq:rho}
\end{equation}
where $\lambda\in[0,1]$ is a weight parameter to balance the expected cost and tail risk, and $\alpha\in(0,1)$ represents the confidence level of CVaR. 
Notice that this risk measure is more general than CVaR and expectation because it has CVaR or expectation as a special case when $\lambda=1$ or $\lambda=0$, respectively.

\begin{figure}[ht!]
    \centering    \includegraphics[width=0.4\textwidth]{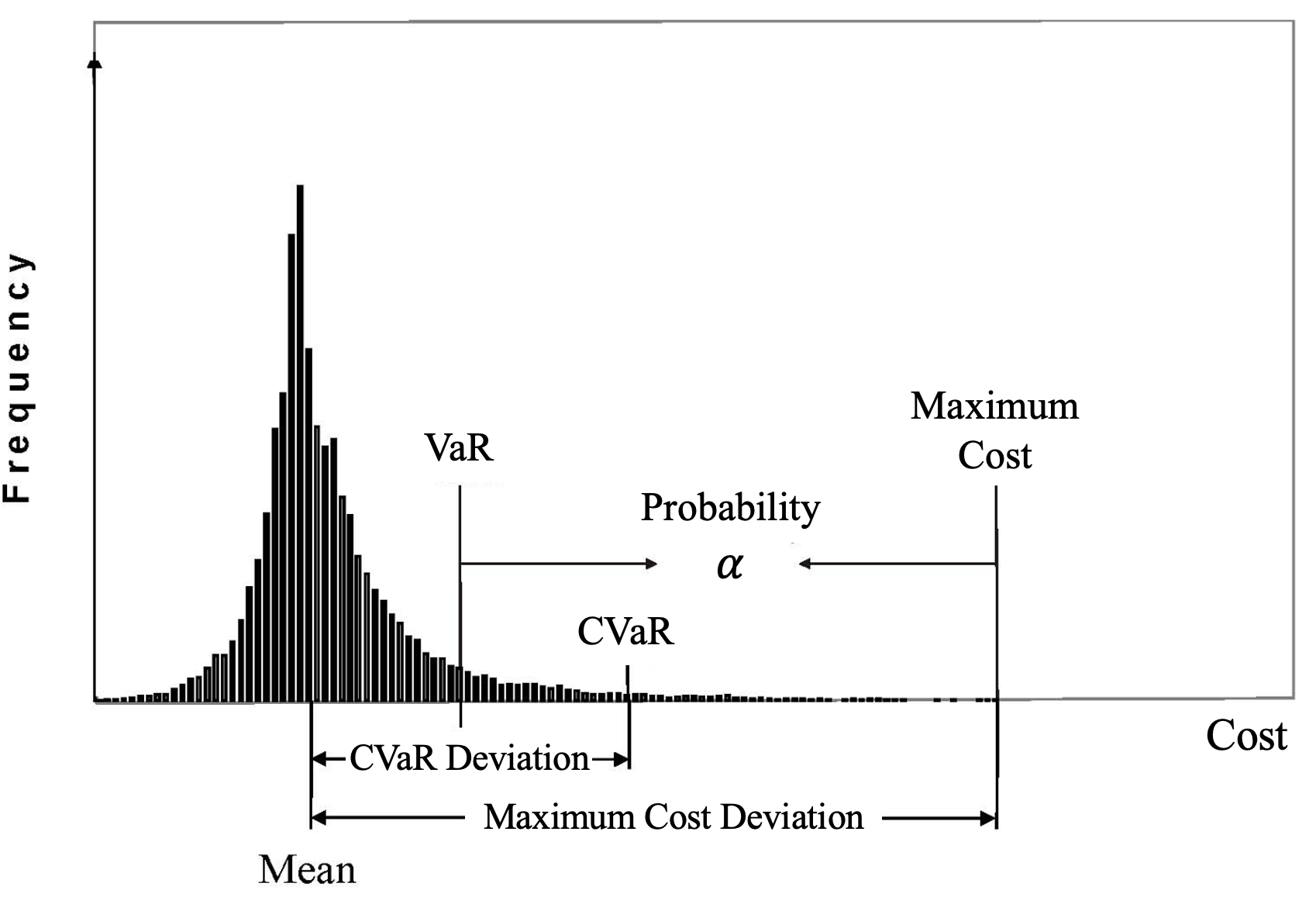}
    \caption{Illustration of CVaR.}
    \label{fig:cvar}
\end{figure}

Following the results by \citet{rockafellar2002conditional}, the upper $\alpha$-tail CVaR can be expressed as the optimization problem below:
\begin{equation}
\text{CVaR}_{\alpha}[c_t|s_{[1,t-1]}]:=\min_{\eta_t\in\mathbb{R}}\left\lbrace \eta_t+\frac{1}{\alpha}\mathbb{E}[[c_t-\eta_t]_{+}|s_{[1,t-1]}]\right\rbrace,\label{eq:cvar}
\end{equation}
where $[a]_{+}:=\max\{a,0\}$, and {\color{black}$\eta_t$ is an auxiliary decision variable to learn the tail distribution. The optimal $\eta$ is attained at $\eta_t^* = \text{VaR}_{\alpha}[c_t|s_{[1,t-1]}] := \inf\{v: \mathbb{P}(c_t\le v)\ge 1-\alpha\}$, which is helpful to calculate CVaR (mean of the upper $\alpha$-tail distribution $\mathbb{E}[c_t|c_t>\eta_t^*]$).}  
Please see Figure \ref{fig:cvar} for an illustration of the CVaR measure. 
Selecting a small $\alpha$ value makes CVaR sensitive to rare but high costs. Because $c_t\in[0,1]$, we restrict the $\eta_t$-variable to be within $[0,1]$ for all $t\ge 1$. 

\subsection{Expected Conditional Risk Measures}
We consider a class of multi-period risk function $\mathbb{F}$ mapping from $\mathcal{Z}_{1,\infty}:=\mathcal{Z}_1\times \mathcal{Z}_2\times\cdots$ to $\mathbb{R}$ below:
\begin{align}
\mathbb{F}(c_{[1,\infty]}|s_1) = c_1+\gamma\varrho^{s_1}_2(c_2)+\lim_{T\to\infty}\sum_{t=3}^T\gamma^{t-1}\mathbb{E}_{s_{[1,t-1]}}\left[{\varrho_t^{s_{[1,t-1]}}}(c_t)\right]\label{ECRMs}, 
\end{align}
where ${\varrho_t^{s_{[1,t-1]}}}$ is the coherent one-step {conditional} risk measure mapping from $\mathcal{Z}_t$ to $\mathcal{Z}_{t-1}$ defined in Eq.\ \eqref{eq:rho} to represent the risk given the information available up to stage $t-1$, and the expectation is taken with respect to the random history $s_{[1,t-1]}$. 
This class of multi-period risk measures is called expected conditional risk measures (ECRMs) \citep{homem2016risk}. 

Using the specific risk measure defined in \eqref{eq:rho} and \eqref{eq:cvar} and applying tower property of expectations on \eqref{ECRMs}, we have
\begin{align}
\min_{a_{[1,\infty]}}\mathbb{F}(c_{[1,\infty]}|s_1)
    &=\min_{a_1,\eta_2}\Big\{C(s_1,a_1)+\gamma\lambda\eta_2\nonumber\\
    &+\gamma\mathbb{E}^{s_1}_{s_2}\Big[\min_{a_2,\eta_3}\Big\{\frac{\lambda}{\alpha}[C(s_2,a_2)-\eta_2]_++(1-\lambda)C(s_2,a_2)+\gamma\lambda\eta_3\nonumber\\
   & +\gamma\mathbb{E}^{s_2}_{s_3}\Big[\min_{a_3,\eta_4}\Big\{\frac{\lambda}{\alpha}[C(s_3,a_3)-\eta_3]_++(1-\lambda)C(s_3,a_3)+\gamma\lambda\eta_4+\cdots\Big\}\Big]\Big\}\Big]\Big\},\label{eq:nested2}
\end{align}
where $\mathbb{E}_{s_{t}}^{s_{t-1}}=\mathbb{E}_{s_{t}}[\cdot|{s_{t-1}}]$ is the conditional expectation and we apply the Markov property to recast $\mathbb{E}_{s_{t}}^{s_{[1,t-1]}}$ as $\mathbb{E}_{s_{t}}^{s_{t-1}}$. The auxiliary variable $\eta_t$ from Eq.\ \eqref{eq:cvar} is decided before taking conditional expectation $\mathbb{E}^{s_{t-1}}_{s_t}$ and thus it should be regarded as a $(t-1)$-stage action, similar to $a_{t-1}$. Here, the optimal $\eta_t$ represents the tail information of state $s_t$'s immediate cost (i.e., $\eta_t^*=\text{VaR}_{\alpha}[C(s_t,a_t)|s_{[1,t-1]}]$), which accounts for the risk when making decisions. We refer interested readers to \citet{yu2022risk} for discussions on the time-consistency of ECRMs and contractive property of the corresponding risk-averse Bellman operator.

Based on our risk-averse formulation \eqref{eq:nested2}, we summarize the key differences compared to a risk-neutral RL as follows. First, we extend the action space $a_t\in\mathcal{A}$ to $(a_t,\eta_{t+1})\in\mathcal{A}\times[0,1]$ for all time steps $t\ge 1$ to include action $\eta_{t+1}$ for learning the tail distribution. Second, we extend the state space $s_t\in\mathcal{S}$ to $(s_t,\eta_{t})\in\mathcal{S}\times[0,1]$ for all $t\ge2$ to record the previous action $\eta_t$. Third, we manipulate the immediate costs by replacing the first-step cost $C(s_1,a_1)$ with $\bar{C}_1(s_1,a_1,\eta_2)=C(s_1,a_1)+\gamma\lambda\eta_{2}$ and replacing $C(s_t,a_t)$ with $\bar{C}(s_t,\eta_t,a_t,\eta_{t+1})=\frac{\lambda}{\alpha}[C(s_t,a_t)-\eta_t]_++(1-\lambda)C(s_t,a_t)+\gamma\lambda\eta_{t+1}$ for $t\ge 2$. Note that for time steps $t\ge 2$, the calculations of $\bar{C}(s_t,\eta_t,a_t,\eta_{t+1})$ involve both the action $\eta_t$ from the previous time step $t-1$ (regarded as part of the current state variable) and the action $\eta_{t+1}$ from the current time step $t$. The inconsistency in defining immediate costs between the first step and others is rooted in the fact that $\eta_t$ is a $(t-1)$-stage action, which must be decided before taking expectation with respect to $s_t$. Due to the difference in immediate costs, our risk-averse formulation \eqref{eq:nested2} cannot be reduced to a risk-neutral RL, and the conventional Bellman equation used in risk-neutral RL cannot be applied here. Therefore, we need to develop new global convergence analyses for risk-averse PG algorithms, while differentiating between the first step and the subsequent ones. 

Since we consider a tabular case for deriving global convergence guarantees, we discretize the $\eta$-space ($[0,1]$) to be $\mathcal{H}=\{\frac{i}{I}, \ i=0,1,\ldots, I\}$. This step can be done exactly if the immediate cost $C(s_t,a_t)$ has finitely many possible values. Otherwise, the discretization becomes finer as we increase $I$. We first provide an optimality guarantee for the discretized problem with finite support $I$ below. The proof is presented in Appendix \ref{append-proof-pre}.

\begin{proposition}[{\color{black}$\epsilon_{opt}$-optimal Discretization}]\label{prop:eta-discretization}
    Denote the ECRM objective function under the original $\eta$-space and the discretized $\mathcal{H}$ space as $\mathbb{F}(c_{[1,\infty]}|s_1)$ and $\mathbb{F}^I(c_{[1,\infty]}|s_1)$, respectively. Then for any given $\epsilon_{opt}> 0$, we have 
    $$\left|\min_{a_{[1,\infty]}}\mathbb{F}(c_{[1,\infty]}|s_1)-\min_{a_{[1,\infty]}}\mathbb{F}^I(c_{[1,\infty]}|s_1)\right|\le \epsilon_{opt}$$
    whenever $I\ge(1+\frac{1}{\alpha})\frac{ \lambda\gamma}{1-\gamma}\frac{1}{\epsilon_{opt}}$.
\end{proposition}
In the sequel, we will focus on the discretized problem where $\eta\in[0,1]$ is replaced with $\eta\in\mathcal{H}$. {\color{black}According to Proposition \ref{prop:eta-discretization}, this discretized problem can find an $\epsilon_{opt}$-optimal policy for the original problem whenever $I$ is sufficiently large. As we will show later, the iteration complexity of the proposed NPG algorithm almost does not depend on the dimension of the state and action space, and thus the discretization resolution will not create further computational burden.}

Because of the differences between the modified immediate costs $\bar{C}_1(s_1,a_1,\eta_2)$ and $\bar{C}(s_t,\eta_t,a_t,\eta_{t+1})$, we should distinguish the value functions and policies for ECRMs-based objectives between the first time step and subsequent ones. Moreover, starting from time step $t\ge2$, problem \eqref{eq:nested2} reduces to a risk-neutral RL with the same form of immediate costs $\bar{C}(s_t,\eta_t,a_t,\eta_{t+1})$, and according to \cite{puterman2014markov}, there exists a deterministic stationary Markov optimal policy. {\color{black}Without loss of optimality}, we consider a class of policies $\pi^{\theta}=(\pi_1^{\theta_1},\pi_2^{\theta_2})\in\Delta(\mathcal{A}\times\mathcal{H})^{|\mathcal{S}|+|\mathcal{S}||\mathcal{H}|}$ where $\pi_1^{\theta_1}(a_1,\eta_2|s_1)$ is the policy for the first time step parameterized by $\theta_1$ and $\pi_2^{\theta_2}(a_t,\eta_{t+1}|s_t,\eta_t)$ is the stationary policy for the following time steps $t\ge 2$ parameterized by $\theta_2$. {\color{black}For ease of presentation, we omit the dependence of $\pi$ on $\theta$ and suppress its arguments when they are clear from the context.}
The goal is to solve the following ECRMs-based optimization problem 
\begin{align}
    \min_{\pi\in{\Delta}(\mathcal{A}\times\mathcal{H})^{|\mathcal{S}|+|\mathcal{S}||\mathcal{H}|}}J^{\pi}(\rho), \label{eq:goal}
\end{align} 
where we denote $\pi^*$ as the optimal policy and $J^*(\rho)$ as the optimal objective value of Model \eqref{eq:goal}, and the value function is defined as
\allowdisplaybreaks
\begin{align*}
&J^{\pi}(\rho) = \mathbb{E}_{s_1\sim\rho}\mathbb{E}_{(a_1,\eta_2)\sim\pi_1(\cdot,\cdot|s_1)}\Big[C(s_1,a_1)+\gamma\lambda\eta_2+\gamma\mathbb{E}^{s_1,a_1}_{s_2}\mathbb{E}_{(a_2,\eta_3)\sim\pi_2(\cdot,\cdot|s_2,\eta_2)}\Big[\frac{\lambda}{\alpha}[C(s_2,a_2)-\eta_2]_+\\
    &+(1-\lambda)C(s_2,a_2)+\gamma\lambda\eta_3+\gamma\mathbb{E}^{s_2,a_2}_{s_3}\mathbb{E}_{(a_3,\eta_4)\sim\pi_2(\cdot,\cdot|s_3,\eta_3)}\Big[\Big\{\frac{\lambda}{\alpha}[C(s_3,a_3)-\eta_3]_+\\
    &+(1-\lambda)C(s_3,a_3)+\gamma\lambda\eta_4+\cdots\Big]\Big]\Big].
\end{align*}

{\color{black}We formalize the above reasoning in the following theorem. 
\begin{theorem}
Denoting the class of all admissible (possibly history-dependent and nonstationary) policies as $\Pi$, we have
    \begin{align*}
    \min_{\pi\in\Pi}J^{\pi}(\rho)=\min_{\pi\in{\Delta}(\mathcal{A}\times\mathcal{H})^{|\mathcal{S}|+|\mathcal{S}||\mathcal{H}|}}J^{\pi}(\rho).
\end{align*} 
\end{theorem}}

\section{Global Convergence of Risk-Averse Natural Policy Gradient Algorithms}
 \label{sec:natural}
In this section, we consider risk-averse NPG algorithms with an entropy-regularized objective function $\min_{\pi} J_{\tau}^{\pi}(\rho):=J^{\pi}(\rho)+\tau \mathcal{R}(\rho,\pi)$.
    Here, $\tau\ge 0$ denotes the regularization parameter and $\mathcal{R}(\rho,\pi)$ is the discounted entropy defined as:
    \begin{align*}
        &\mathcal{R}(\rho,\pi) = \mathbb{E}_{s_1\sim\rho}[\mathcal{R}(s_1,\pi)]\\
=&\mathbb{E}_{\substack{s_1\sim\rho\\a_1,\eta_2\sim\pi_1(\cdot,\cdot|s_1)}}\left[\log\pi_1(a_1,\eta_2|s_1)\right]+\mathbb{E}_{\substack{s_1\sim\rho\\a_1,\eta_2\sim\pi_1(\cdot,\cdot|s_1)\\s_{t}\sim P(\cdot|s_{t-1},a_{t-1})\\(a_t,\eta_{t+1})\sim\pi_2(\cdot,\cdot|s_t,\eta_t)\\t\ge 2}}\left[\sum_{t=2}^{\infty}\gamma^{t-1}\log\pi_2(a_t,\eta_{t+1}|s_t,\eta_t)\right]\\
=&\mathbb{E}_{s_1\sim\rho}\left[\sum_{a_1,\eta_2}\pi_1(a_1,\eta_2|s_1)\log\pi_1(a_1,\eta_2|s_1)\right]\\
&+\frac{\gamma}{1-\gamma}\mathbb{E}_{(s_t,\eta_t)\sim d_{\rho^{\pi}}^{\pi}}\left[\sum_{a_t,\eta_{t+1}}\pi_2(a_t,\eta_{t+1}|s_t,\eta_t)\log\pi_2(a_t,\eta_{t+1}|s_t,\eta_t)\right],
    \end{align*}
    where $\rho$ is the distribution for the initial state $s_1$, and $\rho^{\pi}(s,\eta):=\sum_{s_1}\rho(s_1)\text{Pr}^{\pi_1}(s_2=s,\eta_2=\eta|s_1)$ is the distribution for step-2 state-action pair $s_2,\eta_2$.  
    {\color{black}Define the discounted state visitation distribution when starting from state $s_2,\eta_2$ as $d_{s_2,\eta_2}^{\pi^{\theta}}(s,\eta):=(1-\gamma)\sum_{t=0}^{\infty}\gamma^{t}\text{Pr}^{\pi^{\theta}}(s_{t+2}=s,\eta_{t+2}=\eta|s_2,\eta_2)$ and when starting from initial state distribution $\rho^{\pi}$ as $d_{\rho^{\pi}}^{\pi}(s,\eta):=\mathbb{E}_{(s_2,\eta_2)\sim \rho^{\pi}}[d_{s_2,\eta_2}^{\pi}(s,\eta)]$.} According to the elementary entropy bound, we have $0\le \sum_{x\in\mathcal{X}}p(x)\log\frac{1}{p(x)}\le \log|\mathcal{X}|$, and as a result, $0\ge\mathcal{R}(\rho,\pi)\ge -\log(|\mathcal{A}||\mathcal{H}|)-\frac{\gamma}{1-\gamma}\log(|\mathcal{A}||\mathcal{H}|)=-\frac{1}{1-\gamma}\log(|\mathcal{A}||\mathcal{H}|)$. When $\pi_1(a_1,\eta_2|s_1)$ approaches 0 or 1, we have $\pi_1(a_1,\eta_2|s_1)\log\pi_1(a_1,\eta_2|s_1)\to 0$. On the other hand, when $0<\pi_1(a_1,\eta_2|s_1)<1$, we have $\pi_1(a_1,\eta_2|s_1)\log\pi_1(a_1,\eta_2|s_1)<0$. The same reasoning applies to $\pi_2(a_t,\eta_{t+1}|s_t,\eta_t)$. Since we aim to minimize $J_{\tau}^{\pi}(\rho)$, we discourage premature convergence to near deterministic policies by minimizing $\mathcal{R}(\rho,\pi)$.
    
    To provide a global convergence guarantee for the risk-averse NPG algorithm on this entropy-regularized objective function $J_{\tau}^{\pi}(\rho)$, let us first define the regularized $Q$-functions (also known as soft $Q$-functions) and regularized value functions (also known as soft value functions) for the first time step and the subsequent ones (denoted by $\ \widehat{\cdot}\ $) as follows:
    \begin{subequations}
    \begin{align}
        &J^{\pi}_{\tau}(s_1)=\mathbb{E}_{a_1,\eta_2\sim\pi_1(\cdot,\cdot|s_1)}[\tau\log\pi_1(a_1,\eta_2|s_1)+Q^{\pi}_{\tau}(s_1,a_1,\eta_2)],\label{eq:natural-J1}\\
        &Q_{\tau}^{\pi}(s_1,a_1,\eta_2)=\bar{C}_1(s_1,a_1,\eta_2)+\gamma\mathbb{E}_{s_2\sim P(\cdot|s_1,a_1)}[\widehat{J}^{\pi}_{\tau}(s_2,\eta_2)],\label{eq:natural-Q1}\\
        &\widehat{J}^{\pi}_{\tau}(s_t,\eta_t)=\mathbb{E}_{(a_t,\eta_{t+1})\sim\pi_2(\cdot,\cdot|s_t,\eta_t)}[\tau\log\pi_2(a_t,\eta_{t+1}|s_t,\eta_t)+\widehat{Q}^{\pi}_{\tau}(s_t,\eta_t,a_t,\eta_{t+1})],\ \forall t\ge 2,\label{eq:natural-J2}\\
        &\widehat{Q}^{\pi}_{\tau}(s_t,\eta_t,a_t,\eta_{t+1})=\bar{C}(s_t,\eta_t,a_t,\eta_{t+1})+\gamma\mathbb{E}_{s_{t+1}\sim P(\cdot|s_t,a_t)}[\widehat{J}^{\pi}_{\tau}(s_{t+1},\eta_{t+1})],\ \forall t\ge 2.\label{eq:natural-Q2}
    \end{align}
    \end{subequations}
      Denote the optimal value functions for the entropy-regularized problem as $J^{*}_{\tau}(s_1),\ Q_{\tau}^{*}(s_1,a_1,\eta_2)$, $\widehat{J}^{*}_{\tau}(s_t,\eta_t)$ and $\widehat{Q}^{*}_{\tau}(s_t,\eta_t,a_t,\eta_{t+1})$, and the optimal policy as $\pi_{\tau}^*=(\pi_{\tau,1}^*,\pi_{\tau,2}^*)$, respectively. Similarly, we define the regularized advantage functions for the first time step and the subsequent ones as follows
        \begin{subequations}
    \begin{align}
        &A^{\pi}_{\tau}(s_1,a_1,\eta_2)=J_{\tau}^{\pi}(s_1)-\tau\log\pi_1(a_1,\eta_2|s_1)-Q_{\tau}^{\pi}(s_1,a_1,\eta_2),\label{eq:natural-advantage1}\\
        &\widehat{A}^{\pi}_{\tau}(s_t,\eta_t,a_t,\eta_{t+1})=\widehat{J}^{\pi}_{\tau}(s_t,\eta_t)-\tau\log\pi_2(a_t,\eta_{t+1}|s_t,\eta_t)-\widehat{Q}_{\tau}^{\pi}(s_t,\eta_t,a_t,\eta_{t+1}),\ \forall t\ge 2.\label{eq:natural-advantage2}
    \end{align}
    \end{subequations}
    With the aid of regularized advantage functions, we first derive the gradients of the regularized value function $J_{\tau}^{\pi}(\rho)$ in Theorem \ref{thm:natural-gradient}. All the proofs in this section are presented in Appendix \ref{append-natural}.
    \begin{theorem}[Risk-Averse Policy Gradients with Entropy Regularizer]\label{thm:natural-gradient}
             The gradients of the regularized value function $J_{\tau}^{\pi}(\rho)$ take the following forms:
    \begin{align*}
        \nabla_{\theta_1}J_{\tau}^{\pi}(\rho)=&\mathbb{E}_{s_1\sim\rho}\mathbb{E}_{(a_1,\eta_2)\sim\pi_1(\cdot|s_1)}[\nabla_{\theta_1}\log\pi_1(a_1,\eta_2|s_1)(-A^{\pi}_{\tau}(s_1,a_1,\eta_2))],\\
        \nabla_{\theta_2}J_{\tau}^{\pi}(\rho)=&\frac{\gamma}{1-\gamma}\mathbb{E}_{(s_t,\eta_t)\sim d^{\pi}_{\rho^{\pi}}}\mathbb{E}_{(a_t,\eta_{t+1})\sim\pi_2(\cdot|s_t,\eta_t)}[\nabla_{\theta_2}\log\pi_2(a_t,\eta_{t+1}|s_t,\eta_t)(-\widehat{A}^{\pi}_{\tau}(s_t,\eta_t,a_t,\eta_{t+1}))],
    \end{align*}
    where $\rho^{\pi}(s,\eta)=\sum_{s_1}\rho(s_1)\text{Pr}^{\pi_1}(s_2=s,\eta_2=\eta|s_1)$ and $d^{\pi}_{\rho^{\pi}}(s,\eta)=\mathbb{E}_{(s_2,\eta_2)\sim \rho^{\pi}}[d_{s_2,\eta_2}^{\pi}(s,\eta)]$.
    \end{theorem}
    Different from the risk-neutral setting, to account for the difference in the coefficients of $\nabla_{\theta_1}J_{\tau}^{\pi}(\rho)$ and  $\nabla_{\theta_2}J_{\tau}^{\pi}(\rho)$, we separate the risk-averse NPG updates for $\theta_1$ and $\theta_2$ as follows
        \begin{subequations}\label{eq:NPG-update-rule}
    \begin{align}
        &\theta_1^{(t+1)} :=\theta_1^{(t)} - {\color{black}\beta} (\mathcal{F}^{\theta_1^{(t)}}_{\rho})^{\dagger}\nabla_{\theta_1}J_{\tau}^{\pi}(\rho),   \label{eq:NPG-update-rule1}\\
        &\theta_2^{(t+1)} :=\theta_2^{(t)} - \beta(\mathcal{F}^{\theta_2^{(t)}}_{\rho})^{\dagger}\nabla_{\theta_2}J_{\tau}^{\pi}(\rho),\label{eq:NPG-update-rule2}
        \end{align}
                \end{subequations}
        where $B^{\dagger}$ denotes the Moore-Penrose pseudoinverse of matrix $B$, and $\mathcal{F}^{\theta_1}_{\rho}, \mathcal{F}^{\theta_2}_{\rho}$ are the Fisher information matrices defined below
        \begin{subequations}\label{eq:fisher}
    \begin{align}
&\mathcal{F}^{\theta_1}_{\rho}:={\color{black}\kappa_1}\mathbb{E}_{s_1\sim\rho,(a_1,\eta_2)\sim\pi_1(\cdot|s_1)}[(\nabla_{\theta_1}\log\pi_1(a_1,\eta_2|s_1)(\nabla_{\theta_1}\log\pi_1(a_1,\eta_2|s_1))^{\mathsf T}],\label{eq:fisher1}\\
&\mathcal{F}^{\theta_2}_{\rho}:={\color{black}\kappa_2}\mathbb{E}_{(s_t,\eta_t)\sim d_{\rho^{\pi}}^{\pi},(a_t,\eta_{t+1})\sim\pi_2(\cdot|s_t,\eta_t)}[(\nabla_{\theta_2}\log\pi_2(a_t,\eta_{t+1}|s_t,\eta_t)(\nabla_{\theta_2}\log\pi_2(a_t,\eta_{t+1}|s_t,\eta_t))^{\mathsf T}].\label{eq:fisher2}
    \end{align}
            \end{subequations}
    Note that the different {\color{black}user-defined coefficients $\kappa_1$ and $\kappa_2$ in \eqref{eq:fisher1} and \eqref{eq:fisher2} are helpful to adjust the learning rates between the first step and the following steps}, which is a major change from the risk-neutral NPG algorithm. {\color{black}We state the results under this general setting and derive conditions on $\beta$ and $\kappa_1,\kappa_2$ that lead to linear convergence rates.} It has been recognized that NPG updates try to control the changes between the old and new policies approximately in terms of the KL divergence \citep[see, e.g., Section 7 in][]{schulman2015trust}. The next two lemmas further specify the forms of the NPG updates \eqref{eq:NPG-update-rule} under softmax parameterization.

\begin{lemma}\label{lem:pseudoinverse}
    Under softmax parameterization, the gradient of the regularized value function satisfies
    \begin{subequations}
        \begin{align}
            &[(\mathcal{F}^{\theta_1}_{\rho})^{\dagger}\nabla_{\theta_1}J_{\tau}^{\pi}(\rho)](s,a,\eta)=-{\color{black}\frac{1}{\kappa_1}}A^{\pi}_{\tau}(s,a,\eta)+c(s)\ \text{and} \label{eq:pseudo1}\\
            & [(\mathcal{F}^{\theta_2}_{\rho})^{\dagger}\nabla_{\theta_2}J_{\tau}^{\pi}(\rho)](s_t,\eta_t,a_t,\eta_{t+1})=-{\color{black}\frac{\gamma}{\kappa_2(1-\gamma)}}\widehat{A}^{\pi}_{\tau}(s_t,\eta_t,a_t,\eta_{t+1})+c(s_t,\eta_t),\label{eq:pseudo2}
        \end{align}
        where $c(s)$ and $c(s_t,\eta_t)$ are some functions depending only on $s$ and $s_t,\eta_t$, respectively.
    \end{subequations}
\end{lemma}
        \begin{lemma}\label{lem:NPG-update}
        Under softmax parameterization, the entropy-regularized NPG updates \eqref{eq:NPG-update-rule} satisfy
        \begin{subequations}\label{eq:NPG-update}
        \begin{align}
            &\pi_1^{(t+1)}(a_1,\eta_2|s_1)=\frac{1}{Z^{(t)}_1(s_1)}\left(\pi^{(t)}_1(a_1,\eta_2|s_1)\right)^{{\color{black}1-\frac{\beta\tau}{\kappa_1}}}\exp\left({\color{black}-\frac{\beta}{\kappa_1}} Q_{\tau}^{(t)}(s_1,a_1,\eta_2)\right)\ \text{and}\label{eq:NPG-update1}\\
            &\pi_2^{(t+1)}(a_i,\eta_{i+1}|s_i,\eta_i)=\frac{1}{Z^{(t)}_2(s_i,\eta_i)}\left(\pi^{(t)}_2(a_i,\eta_{i+1}|s_i,\eta_i)\right)^{1-{\color{black}\frac{\beta\tau\gamma}{\kappa_2(1-\gamma)}}}\exp\left({\color{black}-\frac{\beta\gamma}{\kappa_2(1-\gamma)}}\widehat{Q}_{\tau}^{(t)}(s_i,\eta_i,a_i,\eta_{i+1})\right),\label{eq:NPG-update2}
        \end{align}
                \end{subequations}
        where $Z^{(t)}_1(s_1)$ and $Z^{(t)}_2(s_i,\eta_i)$ are two normalization factors.
        \end{lemma}

        \subsection{Risk-Averse NPG Algorithms with Exact Policy Evaluation}\label{sec:NPG-exact}
        In this section, we first study the convergence behavior of entropy-regularized NPG assuming exact policy evaluation in every iteration (i.e., the regularized $Q$-functions $Q_{\tau}^{(t)}$ and $\widehat{Q}_{\tau}^{(t)}$ can be evaluated accurately for all $t$). Later in Section \ref{sec:approximate-NPG}, we will focus on the case when we do not have access to exact policy evaluation and derive convergence results under approximate NPG updates. Next, we first show a performance improvement theorem, which quantifies the differences in $J^{(t)}_{\tau}(s_1)$ and $\widehat{J}^{(t)}_{\tau}(s_2,\eta_2)$ between two consecutive iterations, respectively.
         \begin{theorem}[Performance Improvement]\label{thm:performance-improvement-NPG}
        Suppose that $0< \beta\le\min\{{\color{black}\frac{\kappa_2(1-\gamma)}{\tau\gamma},\frac{\kappa_1}{\tau}}\}$. For any state $s_1$, one has
        \small
        \begin{align}
            &J^{(t)}_{\tau}(s_1)-J^{(t+1)}_{\tau}(s_1)
            =\left((-\tau+{\color{black}\frac{\kappa_1}{\beta}})\right)\text{KL}(\pi_1^{(t+1)}(\cdot|s_1)||\pi_1^{(t)}(\cdot|s_1))+{\color{black}{\color{black}\frac{\kappa_1}{\beta}}}\text{KL}(\pi_1^{(t)}(\cdot|s_1)||\pi_1^{(t+1)}(\cdot|s_1))\nonumber\\
            &+\mathbb{E}_{\substack{a_1,\eta_2\sim\pi_1^{(t+1)}(\cdot|s_1)\\s_2\sim P(\cdot|s_1,a_1)\\(s_i,\eta_i)\sim d_{(s_2,\eta_2)}^{{(t+1)}}}}\left[{\color{black}(-\frac{\tau\gamma}{1-\gamma}+\frac{\kappa_2}{\beta})}\text{KL}(\pi_2^{(t+1)}(\cdot|s_i,\eta_i)||\pi_2^{(t)}(\cdot|s_i,\eta_i))+{\color{black}\frac{\kappa_2}{\beta}}\text{KL}(\pi_2^{(t)}(\cdot|s_i,\eta_i)||\pi_2^{(t+1)}(\cdot|s_i,\eta_i))\right],\label{eq:improvement1}
            \end{align}
            \normalsize
and for any states $s_2,\eta_2$, one has
\small
            \begin{align}
            &\widehat{J}^{(t)}_{\tau}(s_2,\eta_2)-\widehat{J}^{(t+1)}_{\tau}(s_2,\eta_2)\nonumber\\
            =&\mathbb{E}_{(s_i,\eta_i)\sim d_{(s_2,\eta_2)}^{{(t+1)}}}\Big[{\color{black}(-\frac{\tau}{1-\gamma}+\frac{\kappa_2}{\beta\gamma})}\text{KL}(\pi_2^{(t+1)}(\cdot|s_i,\eta_i)||\pi_2^{(t)}(\cdot|s_i,\eta_i))+{\color{black}\frac{\kappa_2}{\beta\gamma}}\text{KL}(\pi_2^{(t)}(\cdot|s_i,\eta_i)||\pi_2^{(t+1)}(\cdot|s_i,\eta_i))\Big].\label{eq:improvement2}
        \end{align}
        \normalsize
        As a result, the regularized value functions are monotonically improving, i.e., $J_{\tau}^{(t)}(s_1)\ge J_{\tau}^{(t+1)}(s_1)$ and $\widehat{J}_{\tau}^{(t)}(s_2,\eta_2)\ge\widehat{J}_{\tau}^{(t+1)}(s_2,\eta_2)$ for all $s_1,s_2,\eta_2$.
    \end{theorem}
    A direct consequence of Theorem \ref{thm:performance-improvement-NPG} is the monotonicity of the soft $Q$-function:
        \begin{align}
        &\widehat{Q}_{\tau}^{(t+1)}(s_i,\eta_i,a_i,\eta_{i+1})=\bar{C}(s_i,\eta_i,a_i,\eta_{i+1})+\gamma\mathbb{E}_{s_{i+1}}[\widehat{J}_{\tau}^{(t+1)}(s_{i+1},\eta_{i+1})]\nonumber\\
        \le& \bar{C}(s_i,\eta_i,a_i,\eta_{i+1})+\gamma\mathbb{E}_{s_{i+1}}[\widehat{J}_{\tau}^{(t)}(s_{i+1},\eta_{i+1})]=\widehat{Q}_{\tau}^{(t)}(s_i,\eta_i,a_i,\eta_{i+1}),\ \forall s_i,\eta_i,a_i,\eta_{i+1}.\label{eq:monotonicity-Q}
    \end{align}
    Using these results, one can show that the risk-averse NPG algorithm enjoys linear convergence rates in terms of the optimal soft $Q$-functions and the associated log policies for both the first time step and subsequent ones, as presented in the following theorem. 
    For notation simplicity, we denote ${\color{black}\omega=1-\frac{\beta\tau\gamma}{\kappa_2(1-\gamma)}}$ where we have $0\le \omega<1$ if $0<\beta\le{\color{black}\frac{\kappa_2(1-\gamma)}{\tau\gamma}}$.
    \begin{theorem}[Linear Convergence of Exact Risk-Averse NPG]\label{thm:linear-NPG}
         For any learning rate {\color{black}$0<\beta\le\min\{\frac{\kappa_2(1-\gamma)}{\tau\gamma},\frac{\kappa_1}{\tau}\}$ and $\frac{\kappa_1}{\kappa_2}<\frac{1}{\gamma}$}, the risk-averse entropy-regularized NPG updates \eqref{eq:NPG-update} satisfy
        \begin{align*}
            &(i): ||\widehat{Q}^*_{\tau}-\widehat{Q}_{\tau}^{(t+1)}||_{\infty}\le C_1\gamma(1-{\color{black}\frac{\beta\tau\gamma}{\kappa_2}})^t,\ \forall t\ge 0,\\
            &(ii): ||\log\pi_{\tau,2}^*-\log\pi_2^{(t+1)}||_{\infty}\le \frac{2C_1}{\tau}(1-{\color{black}\frac{\beta\tau\gamma}{\kappa_2}})^t,\ \forall t\ge 0,\\
            &(iii): ||Q_{\tau}^*-Q_{\tau}^{(t+1)}||_{\infty}\le C_1\gamma(2+\gamma)(1-{\color{black}\frac{\beta\tau\gamma}{\kappa_2}})^t,\ \forall t\ge 0,\\
            &(iv): ||\log\pi_{\tau,1}^*-\log\pi_1^{(t+1)}||_{\infty}\le \frac{2}{\tau}(C_2(1-{\color{black}\frac{\beta\tau\gamma}{\kappa_2}})+C_3)(1-{\color{black}\frac{\beta\tau\gamma}{\kappa_2}})^{t},\ \forall t\ge 0,
        \end{align*}
        where $C_1=||\widehat{Q}_{\tau}^*-\widehat{Q}_{\tau}^{(0)}||_{\infty}+2\omega\tau||\log\pi_2^{(0)}-\log\pi_{\tau,2}^*||_{\infty}$, $C_2=||Q_{\tau}^*+\tau\log\xi_1^{(0)}||_{\infty}$, and $C_3={\color{black}\frac{\gamma(2+\gamma)}{1-\frac{\kappa_1}{\kappa_2}\gamma}}C_1+\frac{\beta\tau}{\kappa_1}||{Q}_{\tau}^*-{Q}_{\tau}^{(0)}||_{\infty}$.
    \end{theorem}
    Note that Theorem \ref{thm:linear-NPG} applies to any learning rate {\color{black}$\beta$ in the range of $(0,\min\{\frac{\kappa_2(1-\gamma)}{\tau\gamma},\frac{\kappa_1}{\tau}\}]$}, including small $\beta$, and {\color{black}$\frac{\kappa_1}{\kappa_2}<\frac{1}{\gamma}$ further controls the learning rate ratio between the first step and the subsequent ones}. From Theorem \ref{thm:linear-NPG}, {\color{black}to reach $||\hat{Q}_{\tau}^*-\hat{Q}_{\tau}^{(t+1)}||_{\infty}\le\epsilon$, the risk-averse NPG method needs no more than $\frac{{\color{black}\kappa_2}}{\beta\tau\gamma}\log(\frac{C_1\gamma}{\epsilon})$ iterations}, and to reach $||Q_{\tau}^*-Q_{\tau}^{(t+1)}||_{\infty}\le\epsilon$, the risk-averse NPG method needs no more than $\frac{{\color{black}\kappa_2}}{\beta\tau\gamma}\log(\frac{C_1\gamma(2+\gamma)}{\epsilon})$ iterations. {\color{black}When $\kappa_2=\gamma$, this iteration complexity reduces to the risk-neutral result in \cite{cen2022fast}.} Note that this iteration complexity bound does not contain any hidden constants and almost does not depend on the dimensions of the MDP (except for very weak dependency in $C_1$).
    \begin{remark}[Linear convergence of soft value functions]\label{remark:linear-NPG-J} From Theorem \ref{thm:linear-NPG}, we can also derive the linear convergence rate of the soft value functions, i.e.,
    \begin{align}
        ||J^*_{\tau}-J_{\tau}^{(t+1)}||_{\infty}\le \Big(2(C_2(1-\frac{\beta\tau\gamma}{\kappa_2})+C_3)+C_1\gamma(2+\gamma)\Big)(1-{\color{black}\frac{\beta\tau\gamma}{\kappa_2}})^t.\label{eq:linear-convergence-J}
    \end{align}
        To see this, we first note that 
        \begin{align}
        J_{\tau}^*(s_1)=\min_{\pi_1\in\Delta}\left\{\sum_{a_1,\eta_2}\pi_1(a_1,\eta_2|s_1)Q_{\tau}^*(s_1,a_1,\eta_2)+\tau\sum_{a_1,\eta_2}\pi_1(a_1,\eta_2|s_1)\log\pi_1(a_1,\eta_2|s_1)\right\},\label{eq:J_tau_1}
        \end{align}
        where $Q_{\tau}^*(s_1,a_1,\eta_2)=Q_{\tau}^{\pi_{\tau,2}^*}(s_1,a_1,\eta_2)$. Then according to Corollary 5 and Eq.~(32) in \cite{nachum2017bridging}, we have
        \begin{align}
            J_{\tau}^*(s_1)=\tau\log\pi_{\tau,1}^*(a_1,\eta_2|s_1)+Q_{\tau}^*(s_1,a_1,\eta_2),\ \forall s_1, a_1,\eta_2, \label{eq:nachum-1}
        \end{align}
        where $\pi_{\tau,1}^*$ is the minimizer of Eq.~\eqref{eq:J_tau_1} and thus is the optimal policy for the first step.
        This implies
    \begin{align}
        &|{J}_{\tau}^*(s_1)-{J}_{\tau}^{(t+1)}(s_1)|\nonumber\\
        =&\left|\mathbb{E}_{a_1,\eta_2\sim\pi_1^{(t+1)}}\left[(\tau\log\pi_{\tau,1}^*(a_1,\eta_2|s_1)+{Q}^*_{\tau}(s_1,a_1,\eta_2))-(\tau\log\pi_1^{(t+1)}(a_1,\eta_2|s_1)+{Q}^{(t+1)}_{\tau}(s_1,a_1,\eta_2))\right]\right|\nonumber\\
        \le & \tau||\log\pi_{\tau,1}^*-\log\pi_1^{(t+1)}||_{\infty}+||{Q}^*_{\tau}-{Q}^{(t+1)}_{\tau}||_{\infty}\nonumber\\
        \le & \Big(2(C_2(1-\frac{\beta\tau\gamma}{\kappa_2})+C_3)+C_1\gamma(2+\gamma)\Big)(1-{\color{black}\frac{\beta\tau\gamma}{\kappa_2}})^t.\label{eq:J*-J-t+1}
    \end{align}
    \end{remark}
\begin{remark}[\small Iteration complexity for achieving an $\epsilon$-optimal policy of the original MDP] The convergence rates established in Theorem \ref{thm:linear-NPG} and Remark \ref{remark:linear-NPG-J} are for achieving the optimal regularized value function $J^*_{\tau}$, instead of the optimal value function $J^*$ of the original MDP. However, by selecting a sufficiently small regularization parameter $\tau$, we can guarantee that $J_{\tau}^*\approx J^*$. Specifically, if we set 
    $\tau = \frac{(1-\gamma)\epsilon}{4\log(|\mathcal{A}||\mathcal{H}|)}$,
    then by Eq.~\eqref{eq:linear-convergence-J}, we can achieve $||J^*_{\tau}-J_{\tau}^{(t+1)}||_{\infty}\le\epsilon/2$ via no more than an order of $\frac{4{\color{black}\kappa_2}\log(|\mathcal{A}||\mathcal{H}|)}{(1-\gamma)\epsilon\beta\gamma}\log(\frac{1}{\epsilon})$ iterations (where we hide the dependencies that are logarithmic on the problem parameters). Recall that the optimal policies to the original and regularized problems are $\pi^*$ and $\pi_{\tau}^*$, respectively. It then follows that
    \begin{align*}
        J^{\pi^{(t+1)}}(s)-J^{\pi^*}(s)=&J^{\pi^{(t+1)}}(s)-J_{\tau}^{\pi^{(t+1)}}(s)+J_{\tau}^{\pi^{(t+1)}}(s)-J_{\tau}^{\pi_{\tau}^*}(s)+J_{\tau}^{\pi_{\tau}^*}(s)-J^{\pi^*}(s)\\
        \le &||J^{\pi^{(t+1)}}(s)-J_{\tau}^{\pi^{(t+1)}}(s)||_{\infty}+||J_{\tau}^{\pi^{(t+1)}}-J_{\tau}^{\pi_{\tau}^*}||_{\infty}+||J_{\tau}^{\pi_{\tau}^*}(s)-J^{\pi^*}(s)||_{\infty}\\
        \le &\frac{2\tau\log(|\mathcal{A}||\mathcal{H}|)}{1-\gamma}+\frac{\epsilon}{2}=\epsilon
    \end{align*}
    where the last inequality uses the fact that, for any policy $\pi$, we have $||J_{\tau}^{\pi}-J^{\pi}||_{\infty}=\tau\max_s|\mathcal{R}(s,\pi)|\le \frac{\tau\log(|\mathcal{A}||\mathcal{H}|)}{1-\gamma}$ and $J^{\pi_{\tau}^*}(s)\ge J^{\pi^*}(s)\ge J_{\tau}^{\pi^*}(s)\ge J_{\tau}^{\pi_{\tau}^*}(s)\ge J^{\pi_{\tau}^*}(s)-\frac{\tau\log(|\mathcal{A}||\mathcal{H}|)}{1-\gamma}$.
\end{remark}

     \proof[Proof of Theorem \ref{thm:linear-NPG}]{
     Recall that $\omega=1-{\color{black}\frac{\beta\tau\gamma}{\kappa_2(1-\gamma)}}\ (0\le\omega<1)$. Following \cite{cen2022fast}, let us define two auxiliary sequences $\{\xi_1^{(t)}\}$ and $\{\xi_2^{(t)}\}$ for the first time step and subsequent ones, respectively, by
     \begin{subequations}\label{eq:auxiliary-xi}
    \begin{align}
        &\xi_1^{(0)}(s,a,\eta):=||\exp{(-Q_{\tau}^*(s,\cdot,\cdot)/\tau)}||_1\pi_1^{(0)}(a,\eta|s),\label{eq:xi_10}\\
        &\xi_2^{(0)}(s,\eta,a,\eta^{\prime})=||\exp{(-\widehat{Q}_{\tau}^*(s,\eta,\cdot,\cdot)/\tau)}||_1\pi_2^{(0)}(a,\eta^{\prime}|s,\eta),\\
        &\xi_1^{(t+1)}(s,a,\eta):=[\xi_1^{(t)}(s,a,\eta)]^{\color{black}1-\frac{\beta\tau}{\kappa_1}}\exp{\left({\color{black}-\frac{\beta}{\kappa_1}}Q_{\tau}^{(t)}(s,a,\eta)\right)},\label{eq:xi_1t}\\
        &\xi_2^{(t+1)}(s,\eta,a,\eta^{\prime}):=[\xi_2^{(t)}(s,\eta,a,\eta^{\prime})]^{\omega}\exp{\left((1-\omega)\frac{-\widehat{Q}_{\tau}^{(t)}(s,\eta,a,\eta^{\prime})}{\tau}\right)}.\label{eq:xi_2}
    \end{align}
    \end{subequations}
    From Eq.~\eqref{eq:auxiliary-xi}, using mathematical induction, we observe $\pi_1^{(t)}(\cdot,\cdot|s)=\frac{\xi_1^{(t)}(s,\cdot,\cdot)}{||\xi_1^{(t)}(s,\cdot,\cdot)||_1}$ and $\pi_2^{(t)}(\cdot,\cdot|s,\eta)=\frac{\xi_2^{(t)}(s,\eta,\cdot,\cdot)}{||\xi_2^{(t)}(s,\eta,\cdot,\cdot)||_1}$.
    It directly follows from Eq.~\eqref{eq:xi_2} that
    \begin{align}
       ||\widehat{Q}_{\tau}^*+\tau\log\xi_2^{(t+1)}||_{\infty}
        =& ||\widehat{Q}_{\tau}^*+\tau\omega\log\xi_2^{(t)}-(1-\omega)\widehat{Q}_{\tau}^{(t)}||_{\infty}\nonumber\\      
        =& ||\omega(\widehat{Q}_{\tau}^*+\tau\log\xi_2^{(t)})+(1-\omega)(\widehat{Q}_{\tau}^*-\widehat{Q}_{\tau}^{(t)})||_{\infty}\nonumber\\
        \le &\omega||\widehat{Q}_{\tau}^*+\tau\log\xi_2^{(t)}||_{\infty}+(1-\omega)||\widehat{Q}_{\tau}^*-\widehat{Q}_{\tau}^{(t)}||_{\infty}.\label{eq:linear-system1}
    \end{align}
    Using a similar reasoning, Eq.~\eqref{eq:xi_1t} gives us 
    \begin{align}
        ||{Q}_{\tau}^*+\tau\log\xi_1^{(t+1)}||_{\infty}
        \le {\color{black}(1-\frac{\beta\tau}{\kappa_1})}||{Q}_{\tau}^*+\tau\log\xi_1^{(t)}||_{\infty}+{\color{black}\frac{\beta\tau}{\kappa_1}}||{Q}_{\tau}^*-{Q}_{\tau}^{(t)}||_{\infty}.\label{eq:Q_tau_t}
    \end{align}
    We first show that $||\widehat{Q}_{\tau}^*-\widehat{Q}_{\tau}^{(t)}||_{\infty}$ can be controlled by an auxiliary sequence $||\widehat{Q}_{\tau}^*+\tau\log\xi_2^{(t)}||_{\infty}$, where the proof mirrors the one for Lemma 3 in \citet{cen2022fast}.
    \begin{lemma}\citep{cen2022fast}\label{lem:NPG-Q}
        For any learning rate $0<\beta\le {\color{black}\frac{\kappa_2(1-\gamma)}{\tau\gamma}}$, the risk-averse entropy-regularized NPG updates \eqref{eq:NPG-update} satisfy
    \begin{align}
        &||\widehat{Q}_{\tau}^*-\widehat{Q}_{\tau}^{(t+1)}||_{\infty}\le \gamma\omega^{t+1}||\widehat{Q}_{\tau}^{(0)}+\tau\log\xi_2^{(0)}||_{\infty}+\gamma||\widehat{Q}_{\tau}^*+\tau\log\xi_2^{(t+1)}||_{\infty}.\label{eq:linear-system2}
    \end{align}
    \end{lemma}

It is then straightforward to combine Eq.~\eqref{eq:linear-system1} and \eqref{eq:linear-system2} in the following linear system
\begin{align}
    x_{t+1}\le Ax_t+\gamma\omega^{t+1}y, \label{eq:linear-system}
\end{align}
where 
\begin{align*}
    A:=\begin{pmatrix}
        \gamma(1-\omega) &\gamma\omega\\
        1-\omega & \omega
    \end{pmatrix},\ 
    x_t:=\begin{pmatrix}
        ||\widehat{Q}_{\tau}^*-\widehat{Q}_{\tau}^{(t)}||_{\infty}\\
        ||\widehat{Q}_{\tau}^*+\tau\log\xi_2^{(t)}||_{\infty}
    \end{pmatrix},\ 
    y:=\begin{pmatrix}
        ||\widehat{Q}_{\tau}^{(0)}+\tau\log\xi_2^{(0)}||_{\infty}\\
        0
    \end{pmatrix}.
\end{align*}

\begin{proposition}\citep{cen2022fast}\label{prop:linear-system}
    Using the linear system \eqref{eq:linear-system}, we obtain for all $t\ge 0$,
    \begin{align}
        &||\widehat{Q}_{\tau}^*-\widehat{Q}_{\tau}^{(t+1)}||_{\infty}\le (1-{\color{black}\frac{\beta\tau\gamma}{\kappa_2}})^t\gamma(||\widehat{Q}_{\tau}^*-\widehat{Q}_{\tau}^0||_{\infty}+2\omega\tau||\log\pi_2^{(0)}-\log\pi_{\tau,2}^*||_{\infty}),\label{eq:hatQ}\\
        &||\widehat{Q}_{\tau}^*+\tau\log\xi_2^{(t+1)}||_{\infty}\le(1-{\color{black}\frac{\beta\tau\gamma}{\kappa_2}})^t(||\widehat{Q}_{\tau}^*-\widehat{Q}_{\tau}^0||_{\infty}+2\omega\tau||\log\pi_2^{(0)}-\log\pi_{\tau,2}^*||_{\infty}).
    \end{align}
\end{proposition}
Eq.~\eqref{eq:hatQ} establishes Assertion (i) in Theorem \ref{thm:linear-NPG}.
{\color{black}Moreover, from Eq.~(11) in \cite{nachum2017bridging}, i.e., 
    \begin{align}
    \widehat{J}_{\tau}^*(s_2,\eta_2)=\tau\log\pi_{\tau,2}^*(a_2,\eta_3|s_2,\eta_2)+\widehat{Q}^*_{\tau}(s_2,\eta_2,a_2,\eta_3),\ \forall s_2,\eta_2,a_2,\eta_3,\label{eq:Nachum}
    \end{align}
    we have 
    \begin{align}
        \pi^*_{\tau,2}(\cdot,\cdot|s,\eta)=\frac{\exp(-\widehat{Q}_{\tau}^*(s,\eta,\cdot,\cdot)/\tau)}{\left\|\exp(-\widehat{Q}_{\tau}^*(s,\eta,\cdot,\cdot)/\tau)\right\|_1}.\label{eq:optimal-pi-nachum}
    \end{align}
    We also have $\pi_2^{(t+1)}(\cdot,\cdot|s,\eta)=\frac{\xi_2^{(t+1)}(s,\eta,\cdot,\cdot)}{||\xi_2^{(t+1)}(s,\eta,\cdot,\cdot)||_1}=\frac{\exp{(\log{\xi_2^{(t+1)}(s,\eta,\cdot,\cdot)})}}{||\exp{(\log{\xi_2^{(t+1)}}(s,\eta,\cdot,\cdot))}||_1}$.
    It then follows from some elementary properties of the softmax function \citep[see, e.g., Appendix A.2 in ][]{cen2022fast} that for all $\theta_1,\theta_2\in\mathbb{R}^{|\mathcal{A}||\mathcal{H}|}$, 
    \begin{align}
    &|\log(||\exp{(\theta_1)}||_1)-\log(||\exp{(\theta_2)}||_1)|\le ||\theta_1-\theta_2||_{\infty}\ \text{and}\label{eq:softmax-property1}\\
     &   ||\log\pi^{\theta_1}-\log\pi^{\theta_2}||_{\infty}\le 2||\theta_1-\theta_2||_{\infty},\label{eq:softmax-property}
    \end{align}
    where $\pi^{\theta}(a,\eta)=\frac{\exp{(\theta_{a,\eta})}}{||\exp{(\theta)}||_1},\ \forall a\in\mathcal{A},\ \eta\in\mathcal{H}$ is the softmax transform of $\theta$.}
     By Eq.~\eqref{eq:softmax-property}, we have
    \begin{align}
        ||\log\pi_{\tau,2}^*-\log\pi_2^{(t+1)}||_{\infty}&\le \frac{2}{\tau}||\widehat{Q}_{\tau}^*+\tau\log\xi_2^{(t+1)}||_{\infty}\nonumber\\
        &\le \frac{2(1-{\color{black}\frac{\beta\tau\gamma}{\kappa_2}})^t}{\tau}(||\widehat{Q}_{\tau}^*-\widehat{Q}_{\tau}^0||_{\infty}+2\omega\tau||\log\pi_2^{(0)}-\log\pi_{\tau,2}^*||_{\infty}).\label{eq:logpi}
    \end{align}
This establishes Assertion (ii) in Theorem \ref{thm:linear-NPG}. {\color{black}Note that although Assertions (i) and (ii) follow from \cite{cen2022fast}, to prove Assertions (iii) and (iv), we need to use the connection between the first time step and the subsequent ones (Eq. \eqref{eq:natural-Q1}) to derive the convergence rates for the first-step value function and policy. }
Specifically, based on the definitions of the soft value and $Q$-functions, we have for all $t\ge 0$,
        \begin{subequations}
        \begin{align}
            &||Q_{\tau}^*-Q_{\tau}^{(t+1)}||_{\infty}\nonumber\\
            =&\max_{s,a,\eta}\left|\gamma\mathbb{E}_{s^{\prime}\sim P(\cdot|s,a)}[\widehat{J}_{\tau}^*(s^{\prime},\eta)-\widehat{J}_{\tau}^{(t+1)}(s^{\prime},\eta)]\right|
            \le \gamma||\widehat{J}_{\tau}^*-\widehat{J}_{\tau}^{(t+1)}||_{\infty}\nonumber\\
            = &\gamma \max_{s,\eta}\left|\mathbb{E}_{a,\eta^{\prime}\sim\pi^{(t)}}[(\tau\log\pi^*_{\tau,2}(a,\eta^{\prime}|s,\eta)+\widehat{Q}^*_{\tau}(s,\eta,a,\eta^{\prime}))-(\tau\log\pi^{(t+1)}_2(a,\eta^{\prime}|s,\eta)+\widehat{Q}^{(t+1)}_{\tau}(s,\eta,a,\eta^{\prime}))]\right|\nonumber\\
            \le&\gamma(\tau||\log\pi_2^{(t+1)}-\log\pi_{\tau,2}^*||_{\infty}+||\widehat{Q}^{(t+1)}_{\tau}-\widehat{Q}^*_{\tau}||_{\infty})\label{eq:Q_t1}\\
            \overset{(a)}{\le}& \gamma(2+\gamma)(1-{\color{black}\frac{\beta\tau\gamma}{\kappa_2}})^{t}(||\widehat{Q}_{\tau}^*-\widehat{Q}_{\tau}^0||_{\infty}+2\omega\tau||\log\pi_2^{(0)}-\log\pi_{\tau,2}^*||_{\infty}),\label{eq:Q_t2}
        \end{align}
                \end{subequations}
        where $(a)$ uses Eq.~\eqref{eq:hatQ} and \eqref{eq:logpi}. This establishes Assertion (iii) in Theorem \ref{thm:linear-NPG}.
        On the other hand, we have
    \begin{align}
        &||{Q}_{\tau}^*+\tau\log\xi_1^{(t+1)}||_{\infty}\nonumber
        \\
        \overset{(a)}{\le}& {\color{black}(1-\frac{\beta\tau}{\kappa_1})}||{Q}_{\tau}^*+\tau\log\xi_1^{(t)}||_{\infty}+\frac{\beta\tau}{\kappa_1}||{Q}_{\tau}^*-{Q}_{\tau}^{(t)}||_{\infty}\nonumber\\
        \overset{(b)}{\le}& {\color{black}(1-\frac{\beta\tau}{\kappa_1})^{t+1}}||{Q}_{\tau}^*+\tau\log\xi_1^{(0)}||_{\infty}+\frac{\beta\tau}{\kappa_1}||{Q}_{\tau}^*-{Q}_{\tau}^{(t)}||_{\infty}+(1-\frac{\beta\tau}{\kappa_1})\frac{\beta\tau}{\kappa_1}||{Q}_{\tau}^*-{Q}_{\tau}^{(t-1)}||_{\infty}\nonumber\\
        &+{\color{black}(1-\frac{\beta\tau}{\kappa_1})^2}\frac{\beta\tau}{\kappa_1}||{Q}_{\tau}^*-{Q}_{\tau}^{(t-2)}||_{\infty}+\cdots+(1-\frac{\beta\tau}{\kappa_1})^t\frac{\beta\tau}{\kappa_1}||{Q}_{\tau}^*-{Q}_{\tau}^{(0)}||_{\infty}\label{eq:xi_1_recursion}\\
        \overset{(c)}{\le} & {\color{black}(1-\frac{\beta\tau}{\kappa_1})^{t+1}}||{Q}_{\tau}^*+\tau\log\xi_1^{(0)}||_{\infty}+(1-\frac{\beta\tau}{\kappa_1})^t\frac{\beta\tau}{\kappa_1}||{Q}_{\tau}^*-{Q}_{\tau}^{(0)}||_{\infty}\nonumber\\
        &+\frac{\gamma(2+\gamma)(1-\frac{\beta\tau\gamma}{\kappa_2})^{t}}{1-\frac{\kappa_1}{\kappa_2}\gamma}(||\widehat{Q}_{\tau}^*-\widehat{Q}_{\tau}^0||_{\infty}+2\omega\tau||\log\pi_2^{(0)}-\log\pi_{\tau,2}^*||_{\infty}),\nonumber\\
        \overset{(d)}{\le} & {\color{black}(1-\frac{\beta\tau\gamma}{\kappa_2})^{t+1}}||{Q}_{\tau}^*+\tau\log\xi_1^{(0)}||_{\infty}+{\color{black}(1-\frac{\beta\tau\gamma}{\kappa_2})^{t}}\frac{\beta\tau}{\kappa_1}||{Q}_{\tau}^*-{Q}_{\tau}^{(0)}||_{\infty}\nonumber\\
        &+\frac{\gamma(2+\gamma)(1-\frac{\beta\tau\gamma}{\kappa_2})^{t}}{1-\frac{\kappa_1}{\kappa_2}\gamma}(||\widehat{Q}_{\tau}^*-\widehat{Q}_{\tau}^0||_{\infty}+2\omega\tau||\log\pi_2^{(0)}-\log\pi_{\tau,2}^*||_{\infty}),\nonumber
    \end{align}
    where $(a)$ is due to Eq.~\eqref{eq:Q_tau_t}, $(b)$ is by recursively applying the inequality $(a)$, $(c)$ uses Eq.~\eqref{eq:Q_t2} and the sum of a geometric series, {\color{black}where the ratio of consecutive terms $\frac{1-\frac{\beta\tau}{\kappa_1}}{1-\frac{\beta\tau\gamma}{\kappa_2}}<1$ if $\frac{\kappa_1}{\kappa_2}<\frac{1}{\gamma}$, and $(d)$ is because $1-\frac{\beta\tau}{\kappa_1}<1-\frac{\beta\tau\gamma}{\kappa_2}$}.
    According to Eq.~\eqref{eq:nachum-1}, we obtain $\pi^*_{\tau,1}(\cdot,\cdot|s_1)\propto \exp(-{Q}_{\tau}^*(s_1,\cdot,\cdot)/\tau)$.
    Because $\pi_1^{(t+1)}(\cdot,\cdot|s_1)\propto \exp(\log\xi_1^{(t+1)}(s,\cdot,\cdot))$, according to Eq.~\eqref{eq:softmax-property}, we have
    \begin{align*}
        &||\log\pi_{\tau,1}^*-\log\pi_1^{(t+1)}||_{\infty}\\
        \le& \frac{2}{\tau}||Q_{\tau}^{*}+\tau\log\xi_1^{(t+1)}||_{\infty}\\
        \le & \frac{2}{\tau} \Big({\color{black}(1-\frac{\beta\tau\gamma}{\kappa_2})^{t+1}}||{Q}_{\tau}^*+\tau\log\xi_1^{(0)}||_{\infty}+{\color{black}(1-\frac{\beta\tau\gamma}{\kappa_2})^{t}}\frac{\beta\tau}{\kappa_1}||{Q}_{\tau}^*-{Q}_{\tau}^{(0)}||_{\infty}\\
        &+\frac{\gamma(2+\gamma)(1-\frac{\beta\tau\gamma}{\kappa_2})^{t}}{1-\frac{\kappa_1}{\kappa_2}\gamma}(||\widehat{Q}_{\tau}^*-\widehat{Q}_{\tau}^0||_{\infty}+2\omega\tau||\log\pi_2^{(0)}-\log\pi_{\tau,2}^*||_{\infty})\Big).
    \end{align*}
    This establishes Assertion (iv) in Theorem \ref{thm:linear-NPG} with general learning rates.
    $\blacksquare$}

 \subsection{Approximate Risk-Averse NPG Algorithms with Inexact Policy Evaluation}\label{sec:approximate-NPG}
 In this section, we focus on the convergence properties of the risk-averse NPG algorithms when the soft $Q$-function is available only in an approximated fashion, e.g., when the value function has to be evaluated using finite samples. Under this setting, at each iteration, given the current policy $\pi^{(t)}$, we do not have access to the exact regularized $Q$-functions ${Q}_{\tau}^{(t)}$ and $\widehat{Q}_{\tau}^{(t)}$. Instead, we use approximate $Q$-functions $\widetilde{Q}_{\tau}^{(t)}$ and $\widetilde{\widehat{Q}}_{\tau}^{(t)}$, with $||\widetilde{Q}_{\tau}^{(t)}-{Q}_{\tau}^{(t)}||_{\infty}\le \delta$ and $||\widetilde{\widehat{{Q}}}_{\tau}^{(t)}-\widehat{Q}_{\tau}^{(t)}||_{\infty}\le \delta$, to update our policy in the first time step and subsequent ones in the following:
 \begin{subequations}\label{eq:inexact-NPG-update}
    \begin{align}
       & \pi_1^{(t+1)}(\cdot,\cdot|s)=\frac{1}{\widetilde{Z}_1^{(t)}(s)}(\pi_1^{(t)}(\cdot,\cdot|s))^{1-{\color{black}\frac{\beta\tau}{\kappa_1}}}\exp\left(-{\color{black}\frac{\beta}{\kappa_1}}\widetilde{Q}_{\tau}^{(t)}(s,\cdot,\cdot)\right),\\
       & \pi_2^{(t+1)}(\cdot,\cdot|s,\eta)=\frac{1}{\widetilde{Z}_2^{(t)}(s,\eta)} (\pi_2^{(t)}(\cdot,\cdot|s,\eta))^{1-{\color{black}\frac{\beta\tau\gamma}{\kappa_2(1-\gamma)}}}\exp\left(-{\color{black}\frac{\beta\gamma}{\kappa_2(1-\gamma)}}\widetilde{\widehat{Q}}_{\tau}^{(t)}(s,\eta,\cdot,\cdot)\right),
    \end{align}
    \end{subequations}
    respectively, where $\widetilde{Z}_1^{(t)}(s)$ and $\widetilde{Z}_2^{(t)}(s,\eta)$ are two normalization factors. Using $J_{\tau}^{(t)}$ and $\widehat{J}_{\tau}^{(t)}$ to denote the exact regularized value functions under policy $\pi^{(t)}$ in the first time step and subsequent ones, respectively, we first bound the differences when evaluating these value functions between two consecutive iterations in the next lemma.
    \begin{lemma}\label{lem:approximate-NPG-improvement}
        Suppose that $0<\beta\le {\color{black}\min\{\frac{\kappa_2(1-\gamma)}{\tau\gamma},\frac{\kappa_1}{\tau}\}}$. Using the update rule in Eq.~\eqref{eq:inexact-NPG-update}, for any state $s_1$, one has 
            $J_{\tau}^{(t)}(s_1)\ge {J}_{\tau}^{(t+1)}(s_1)-\frac{2\gamma}{1-\gamma}||\widetilde{\widehat{{Q}}}_{\tau}^{(t)}-\widehat{Q}_{\tau}^{(t)}||_{\infty}-2||\widetilde{{Q}}_{\tau}^{(t)}-{{Q}}_{\tau}^{(t)}||_{\infty}$.
        For any states $s_2,\eta_2$, one has $\widehat{J}_{\tau}^{(t)}(s_2,\eta_2)\ge \widehat{J}_{\tau}^{(t+1)}(s_2,\eta_2)-\frac{2}{1-\gamma}||\widetilde{\widehat{{Q}}}_{\tau}^{(t)}-\widehat{Q}_{\tau}^{(t)}||_{\infty}$.
    \end{lemma}
    Note that Lemma \ref{lem:approximate-NPG-improvement} is a relaxation of Theorem \ref{thm:performance-improvement-NPG} with some additional terms quantifying the effect of the approximate error. By repeating the argument \eqref{eq:monotonicity-Q} and applying the assumption $||\widetilde{\widehat{{Q}}}_{\tau}^{(t)}-\widehat{Q}_{\tau}^{(t)}||_{\infty}\le \delta$, we reveal the difference between the soft $Q$-function estimates in two consecutive iterations as follows: for any states $s_i,\eta_i,a_i,\eta_{i+1}$, we have
    \begin{align}
        &\widehat{Q}_{\tau}^{(t)}(s_i,\eta_i,a_i,\eta_{i+1})-\widehat{Q}_{\tau}^{(t+1)}(s_i,\eta_i,a_i,\eta_{i+1})\nonumber\\
        =&\gamma\mathbb{E}_{s_{i+1}\sim P(\cdot|s_i,a_i)}[\widehat{J}_{\tau}^{(t)}(s_{i+1},\eta_{i+1})-\widehat{J}_{\tau}^{(t+1)}(s_{i+1},\eta_{i+1})]\nonumber\\
        \ge & -\frac{2\gamma}{1-\gamma}||\widetilde{\widehat{{Q}}}_{\tau}^{(t)}-\widehat{Q}_{\tau}^{(t)}||_{\infty}
        \ge -\frac{2\delta\gamma}{1-\gamma}.\label{eq:approximate-Q-improvement}
    \end{align}
    We then define two auxiliary sequences $\{\widetilde{\xi}_1^{(t)}\}$ and $\{\widetilde{\xi}_2^{(t)}\}$ recursively by
    \begin{subequations}\label{eq:approximate-auxi}
     \begin{align}
        &\widetilde{\xi}_1^{(0)}(s,a,\eta):=||\exp{(-Q_{\tau}^*(s,\cdot,\cdot)/\tau)}||_1\pi_1^{(0)}(a,\eta|s),\\
        &\widetilde{\xi}_2^{(0)}(s,\eta,a,\eta^{\prime})=||\exp{(-\widehat{Q}_{\tau}^*(s,\eta,\cdot,\cdot)/\tau)}||_1\pi_2^{(0)}(a,\eta^{\prime}|s,\eta),\\
        &\widetilde{\xi}_1^{(t+1)}(s,a,\eta):=[\widetilde{\xi}_1^{(t)}(s,a,\eta)]^{1-{\color{black}\frac{\beta\tau}{\kappa_1}}}\exp{\left(-{\color{black}\frac{\beta}{\kappa_1}}\widetilde{Q}_{\tau}^{(t)}(s,a,\eta)\right)},\\
        &\widetilde{\xi}_2^{(t+1)}(s,\eta,a,\eta^{\prime}):=[\widetilde{\xi}_2^{(t)}(s,\eta,a,\eta^{\prime})]^{\omega}\exp{\left((1-\omega)\frac{-\widetilde{\widehat{{Q}}}_{\tau}^{(t)}(s,\eta,a,\eta^{\prime})}{\tau}\right)},\label{eq:approximate-xi_2t}
    \end{align}
    \end{subequations}
    where $\omega=1-{\color{black}\frac{\beta\tau\gamma}{\kappa_2(1-\gamma)}}$.
     From Eq.~\eqref{eq:approximate-auxi}, using mathematical induction, we have $\pi_1^{(t)}(\cdot,\cdot|s)=\frac{\widetilde{\xi}^{(t)}_1(s,\cdot,\cdot)}{||\widetilde{\xi}^{(t)}_1(s,\cdot,\cdot)||_1}$ and $\pi_2^{(t)}(\cdot,\cdot|s,\eta)=\frac{\widetilde{\xi}^{(t)}_2(s,\eta,\cdot,\cdot)}{||\widetilde{\xi}^{(t)}_2(s,\eta,\cdot,\cdot)||_1}$.    Using Eq.~\eqref{eq:approximate-Q-improvement} and the two auxiliary sequences $\{\widetilde{\xi}_1^{(t)}\}$ and $\{\widetilde{\xi}_2^{(t)}\}$, we construct a linear system to track the error dynamics of the policy updates while taking into account inexact policy evaluation in the following lemma.
    \begin{lemma}\citep{cen2022fast}\label{lem:approximate-NPG-linear-system}
        The following linear system tracks the error dynamics of the approximate policy updates:
        \begin{align}
        z_{t+1}\le Bz_t+b, \label{eq:approximate-recursion}
        \end{align}
        where 
        \begin{align*}
    B:=
    \begin{pmatrix}
        \gamma(1-\omega) & \gamma\omega & \gamma\omega\\
        1-\omega & \omega & 0\\
        0 & 0 & \omega
    \end{pmatrix},\   b:=(1-\omega)\delta
    \begin{pmatrix}
        \gamma(2+{\color{black}\frac{2\kappa_2}{\beta\tau}})\\
        1\\
        1+{\color{black}\frac{2\kappa_2}{\beta\tau}}
    \end{pmatrix}\\
    z_t:=
    \begin{pmatrix}
        ||\widehat{Q}^*_{\tau}-\widehat{Q}_{\tau}^{(t)}||_{\infty}\\
        ||\widehat{Q}^*_{\tau}+\tau\log\widetilde{\xi}_2^{(t)}||_{\infty}\\
        \max_{s,\eta,a,\eta^{\prime}}(\widehat{Q}_{\tau}^{(t)}(s,\eta,a,\eta^{\prime})+\tau\log\widetilde{\xi}_2^{(t)}(s,\eta,a,\eta^{\prime}))
    \end{pmatrix}.
     \end{align*}
    \end{lemma}
    Here, matrix $B$ tracks the contraction rate and the term $b$ captures the error introduced by inexact policy evaluation. Using this lemma, we are able to characterize the convergence rate of approximate risk-averse NPG algorithms with inexact policy evaluation in the following theorem. Please refer to Appendix \ref{append-natural} for the proof.
        \begin{theorem}[Linear Convergence of Approximate Risk-Averse NPG]\label{thm:linear-inexact-NPG}
         For any learning rate $0<\beta\le{\color{black}\min\{\frac{\kappa_2(1-\gamma)}{\tau\gamma},\frac{\kappa_1}{\tau}\}}$, the inexact risk-averse NPG updates \eqref{eq:inexact-NPG-update} satisfy
        \begin{align*}
            &(i): ||\widehat{Q}^*_{\tau}-\widehat{Q}_{\tau}^{(t+1)}||_{\infty}\le \gamma\left(C_1(1-{\color{black}\frac{\beta\tau\gamma}{\kappa_2}})^t+C_4\right),\ \forall t\ge 0,\\
            &(ii): ||\log\pi_{\tau,2}^*-\log\pi_2^{(t+1)}||_{\infty}\le \frac{2}{\tau}\left(C_1(1-{\color{black}\frac{\beta\tau\gamma}{\kappa_2}})^t+C_4\right),\ \forall t\ge 0,\\
            &(iii): ||Q_{\tau}^*-Q_{\tau}^{(t+1)}||_{\infty}\le \gamma(2+\gamma)\left(C_1(1-{\color{black}\frac{\beta\tau\gamma}{\kappa_2}})^t+C_4\right),\ \forall t\ge 0,\\
            &(iv): ||\log\pi_{\tau,1}^*-\log\pi_1^{(t+1)}||_{\infty}\le \frac{2}{\tau}\left((C_2(1-{\color{black}\frac{\beta\tau\gamma}{\kappa_2}})+C_3)(1-{\color{black}\frac{\beta\tau\gamma}{\kappa_2}})^{t}+\gamma(2+\gamma)C_4+\delta\right),\ \forall t\ge 0,
        \end{align*}
        where $C_1=||\widehat{Q}_{\tau}^*-\widehat{Q}_{\tau}^0||_{\infty}+2\omega\tau||\log\pi_2^{(0)}-\log\pi_{\tau,2}^*||_{\infty}$, $C_2=||Q_{\tau}^*+\tau\log\xi_1^{(0)}||_{\infty}$, $C_3={\color{black}\frac{\gamma(2+\gamma)}{1-\frac{\kappa_1}{\kappa_2}\gamma}}C_1+\frac{\beta\tau}{\kappa_1}||{Q}_{\tau}^*-{Q}_{\tau}^{(0)}||_{\infty}$, and $C_4=\frac{2\delta}{1-\gamma}(1+{\color{black}\frac{\kappa_2}{\beta\tau}})$.
    \end{theorem}
    Compared to Theorem \ref{thm:linear-inexact-NPG}, Theorem \ref{thm:linear-NPG} is a special case corresponding to $\delta=0$. According to Theorem \ref{thm:linear-inexact-NPG}, if the estimation error in soft $Q$-functions can be upper bounded by $\delta\le \frac{(1-\gamma)\epsilon}{4\gamma(2+\gamma)(1+\frac{{\color{black}\kappa_2}}{\beta\tau})}$,
    then the approximate risk-averse NPG method can reach $||Q_{\tau}^*-Q_{\tau}^{(t+1)}||_{\infty}\le\epsilon$ within $\frac{{\color{black}\kappa_2}}{\beta\tau\gamma}\log(\frac{2C_1\gamma(2+\gamma)}{\epsilon})$ iterations for general learning rates $0<\beta\le{\color{black}\min\{\frac{\kappa_2(1-\gamma)}{\tau\gamma},\frac{\kappa_1}{\tau}\}}$. 
    \begin{remark}[Sample complexity of approximate risk-averse NPG]
        Theorem \ref{thm:linear-inexact-NPG} is useful to derive sample complexity bounds with some known sample complexities for approximate policy evaluation. For example, \cite{li2020breaking} showed that using a generative model, model-based policy evaluation can achieve $||\widetilde{Q}_{\tau}^{\pi}-Q_{\tau}^{\pi}||_{\infty}\le \delta$ for any fixed policy $\pi$ with high probability whenever the number of samples per state-action pair exceeds the order of $\frac{1}{(1-\gamma)^3\delta^2}$ up to some logarithmic factor. From Theorem \ref{thm:linear-inexact-NPG}, the approximate risk-averse NPG algorithm in the SPI case needs at most $\widetilde{O}(\frac{1}{1-\gamma})$ iterations to reach $||Q_{\tau}^*-Q_{\tau}^{(t+1)}||_{\infty}\le\epsilon$, where $\widetilde{O}$ hides any logarithmic factors. Setting $\delta=\frac{(1-\gamma)^2\epsilon}{4\gamma(2+\gamma)}$ and utilizing fresh samples per policy evaluation, we can show that SPI with model-based policy evaluation needs at most $\widetilde{O}(\frac{|\mathcal{S}||\mathcal{A}||\mathcal{H}|}{(1-\gamma)^8\epsilon^2})$ samples to find an $\epsilon$-optimal policy.
    \end{remark}
    
 \section{Numerical Results}
\label{sec:exper}

\begin{figure}[ht!]
    \centering
\subfigure[Average test cost over 10 runs in NPG.]{
	\includegraphics[width=0.45\textwidth]{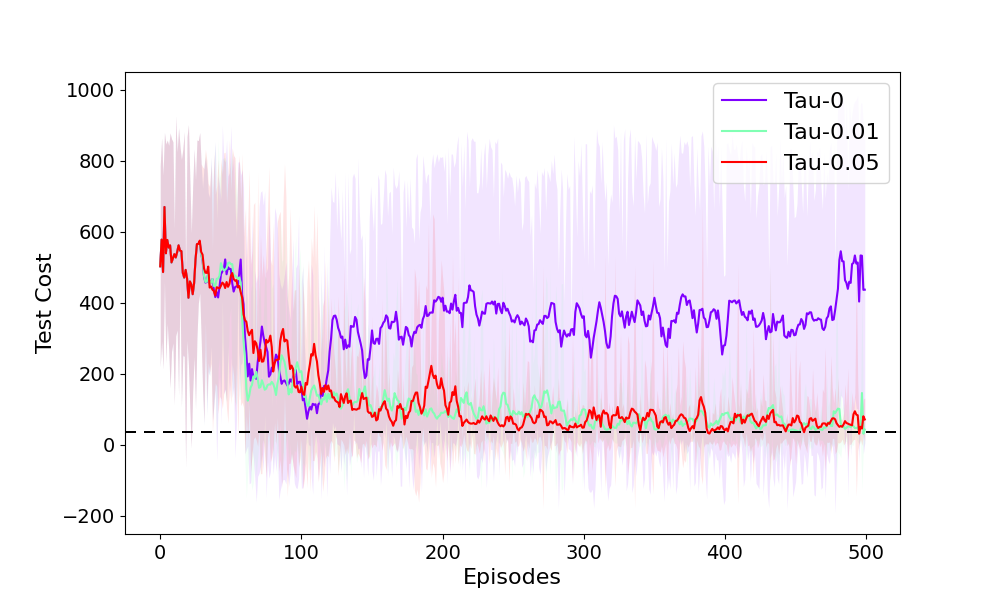}
}
\subfigure[Average test cost over 10 runs in PG.]{
	\includegraphics[width=0.45\textwidth]{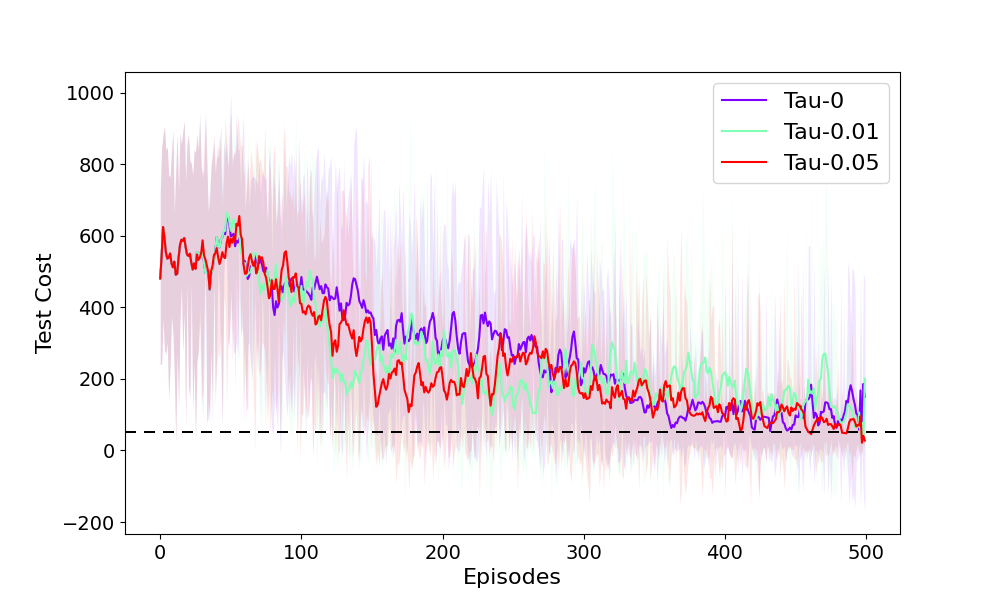}
}
\caption{Risk-averse NPG v.s. PG algorithm with varying $\tau$.}
    \label{fig:results-npg}
\end{figure}

We implement a risk-averse NPG algorithm (Algorithm \ref{alg:npg}) on a $5\times 5$ stochastic Cliffwalk environment, where we utilize neural network approximations for the policy. We compare PG and NPG with varying regularization weight $\tau$ from 0 to 0.05 and present the results in Figure \ref{fig:results-npg}. From Figure \ref{fig:results-npg}, when $\tau=0$, the average test cost of NPG first drops to a desirable level after 100 episodes and then becomes worse over time. This instability of NPG is caused by the numerical issues when computing the inverse of the Fisher information matrix and has also been observed by \cite{kakade2001natural}. As we increase the regularization weight $\tau$, NPG converges to a policy with low cost after 200 episodes, whereas the PG counterpart converges to the same threshold after 500 episodes. {\color{black}More numerical results can be found in our earlier version \citep{yu2023global}.}

\section{Conclusions}
In this paper, we applied a class of dynamic time-consistent coherent risk measures (i.e., ECRMs) on infinite-horizon MDPs and provided a dimension-free linear convergence rate for risk-averse NPG methods with entropy regularization. 
We also considered the case when we cannot evaluate the value functions exactly and derived convergence results under approximate NPG updates.
For future research, it is worth investigating iteration complexities for ECRMs-based PG algorithms with restricted policy classes (e.g., log-linear policy and neural network policy). 


\section*{Acknowledgments}
The work of Xian Yu is supported in part by NSF under grant 2331782. The work of Lei Ying is supported in part by NSF under grants 2112471, 2207548, 2228974, 2240981, and 2331780.


\newpage
\bibliography{Xian_bib}
\bibliographystyle{plainnat}


\newpage

\appendix
\onecolumn

\section{Omitted Proofs in Section \ref{sec:problem}}
\label{append-proof-pre}

\proof{\textbf{of Proposition \ref{prop:eta-discretization}} }[Optimality guarantee for discretization of the $\eta$-space]
Denote the objective function in \eqref{eq:cvar} as $f(\eta):=\eta+\frac{1}{\alpha}\mathbb{E}[[c-\eta]_{+}]$. Clearly, this function is Lipschitz continuous with a Lipschitz constant of $1+\frac{1}{\alpha}$. To see this, we first note that $[c-\eta]_+$ is a Lipschitz continuous function with a Lipschitz constant of $1$. Expectation, scaling, and summation preserve the Lipschitz continuity and we derive the corresponding Lipschitz constant as $1+\frac{1}{\alpha}$.
Denote the optimal solution to $\min_{\eta\in[0,1]}f(\eta)$ as $\eta^*$ and the one to $\min_{\eta\in\mathcal{H}}f(\eta)$ as $\eta^I$. Then we have $|\eta^*-\eta^I|\le \frac{1}{I}$. As a result, $|f(\eta^*)-f(\eta^I)|\le (1+\frac{1}{\alpha})|\eta^*-\eta^I|\le (1+\frac{1}{\alpha})\frac{1}{I}$. Now, denote the ECRM objective function under the original $\eta$-space and the discretized $\mathcal{H}$ space as $\mathbb{F}(c_{[1,\infty]}|s_1)$ and $\mathbb{F}^I(c_{[1,\infty]}|s_1)$, respectively. Then, for any $a_{[1,\infty]}$, we have
\begin{align*}
    \left|\mathbb{F}(c(s_{[1,\infty]},a_{[1,\infty]}))-\mathbb{F}^I(c(s_{[1,\infty]},a_{[1,\infty]}))\right|&\le \gamma\lambda(1+\frac{1}{\alpha})\frac{1}{I}+\gamma^2 \lambda(1+\frac{1}{\alpha})\frac{1}{I}+\gamma^3\lambda(1+\frac{1}{\alpha})\frac{1}{I}+\cdots\\
    &\le \lambda(1+\frac{1}{\alpha})\frac{1}{I}\frac{\gamma}{1-\gamma}\le \epsilon_{opt}
\end{align*}
whenever $I\ge \lambda(1+\frac{1}{\alpha})\frac{1}{\epsilon_{opt}}\frac{\gamma}{1-\gamma}$. Denote $a^*_{[1,\infty]}=\arg\min_{a_{[1,\infty]}}\mathbb{F}(c(s_{[1,\infty]},a_{[1,\infty]})|s_1)$ and $a^I_{[1,\infty]}=\arg\min_{a_{[1,\infty]}}\mathbb{F}^I(c(s_{[1,\infty]},a_{[1,\infty]})|s_1)$. Then
\begin{align*}
    &\left|\min_{a_{[1,\infty]}}\mathbb{F}(c(s_{[1,\infty]},a_{[1,\infty]}))-\min_{a_{[1,\infty]}}\mathbb{F}^I(c(s_{[1,\infty]},a_{[1,\infty]}))\right|\\
    \le & \max\left\{\mathbb{F}(c(s_{[1,\infty]},a^*_{[1,\infty]}))-\mathbb{F}^I(c(s_{[1,\infty]},a^I_{[1,\infty]})), \mathbb{F}^I(c(s_{[1,\infty]},a^I_{[1,\infty]}))-\mathbb{F}(c(s_{[1,\infty]},a^*_{[1,\infty]}))\right\}\\
    \le &\max\left\{\mathbb{F}(c(s_{[1,\infty]},a^I_{[1,\infty]}))-\mathbb{F}^I(c(s_{[1,\infty]},a^I_{[1,\infty]})), \mathbb{F}^I(c(s_{[1,\infty]},a^*_{[1,\infty]}))-\mathbb{F}(c(s_{[1,\infty]},a^*_{[1,\infty]}))\right\}\\
    \le & \epsilon_{opt}
\end{align*}
This completes the proof.

\section{Omitted Proofs in Section \ref{sec:natural}}\label{append-natural}
\proof{\textbf{of Theorem \ref{thm:natural-gradient}} }[Risk-Averse Policy Gradients with Entropy Regularizer]
    According to Eq.~\eqref{eq:natural-J1}, we have
    \begin{align*}
    \nabla_{\theta_1}J^{\pi}_{\tau}(s_1)
    =&\nabla_{\theta_1}(\sum_{a_1,\eta_2}\pi_1(a_1,\eta_2|s_1)(\tau\log\pi_1(a_1,\eta_2|s_1)+Q^{\pi}_{\tau}(s_1,a_1,\eta_2)))\\
    \overset{(a)}{=}&\sum_{a_1,\eta_2}\nabla_{\theta_1}\pi_1(a_1,\eta_2|s_1)(\tau\log\pi_1(a_1,\eta_2|s_1)+Q^{\pi}_{\tau}(s_1,a_1,\eta_2))\\
    &+\sum_{a_1,\eta_2}\pi_1(a_1,\eta_2|s_1)\nabla_{\theta_1}(\tau\log\pi_1(a_1,\eta_2|s_1)+\bar{C}_1(s_1,a_1,\eta_2)+\gamma\mathbb{E}_{s_2}[\widehat{J}^{\pi_2}_{\tau}(s_2,\eta_2)])\\
    \overset{(b)}{=}&\sum_{a_1,\eta_2}\pi_1(a_1,\eta_2|s_1)\nabla_{\theta_1}\log\pi_1(a_1,\eta_2|s_1)(\tau\log\pi_1(a_1,\eta_2|s_1)+Q^{\pi}_{\tau}(s_1,a_1,\eta_2))\\
    &+\gamma\sum_{a_1,\eta_2}\pi_1(a_1,\eta_2|s_1)\sum_{s_2}P(s_2|s_1,a_1)\nabla_{\theta_1}\widehat{J}^{\pi_2}_{\tau}(s_2,\eta_2)\\
    \overset{(c)}{=}&\mathbb{E}_{(a_1,\eta_2)\sim\pi_1}[\nabla_{\theta_1}\log\pi_1(a_1,\eta_2|s_1)(\tau\log\pi_1(a_1,\eta_2|s_1)+Q^{\pi}_{\tau}(s_1,a_1,\eta_2))]\\
    \overset{(d)}{=}&\mathbb{E}_{(a_1,\eta_2)\sim\pi_1}[\nabla_{\theta_1}\log\pi_1(a_1,\eta_2|s_1)(-A^{\pi}_{\tau}(s_1,a_1,\eta_2))]
    \end{align*}
    where $(a)$ is due to Eq.~\eqref{eq:natural-Q1}, $(b)$ is because $\sum_{a_1,\eta_2}\pi_1(a_1,\eta_2|s_1)\nabla_{\theta_1}\tau\log\pi_1(a_1,\eta_2|s_1)=\tau\sum_{a_1,\eta_2}\nabla_{\theta_1}\pi_1(a_1,\eta_2|s_1)=0$, $(c)$ is due to $\nabla_{\theta_1}\widehat{J}^{\pi_2}_{\tau}(s_2,\eta_2)=0$, and $(d)$ is according to \eqref{eq:natural-advantage1} and $\mathbb{E}_{(a_1,\eta_2)\sim\pi_1}[\nabla_{\theta_1}\log\pi_1(a_1,\eta_2|s_1)(-J^{\pi}_{\tau}(s_1))]=0$.
As a result,
\small
    \begin{align*}
    \nabla_{\theta_1}J^{\pi}_{\tau}(\rho) &= \nabla_{\theta_1}\mathbb{E}_{s_1\sim\rho}[J^{\pi}_{\tau}(s_1)]=\mathbb{E}_{s_1\sim\rho}[\nabla_{\theta_1}J^{\pi}_{\tau}(s_1)]=\mathbb{E}_{s_1\sim\rho}\mathbb{E}_{(a_1,\eta_2)\sim\pi_1(\cdot|s_1)}[\nabla_{\theta_1}\log\pi_1(a_1,\eta_2|s_1)(-A^{\pi}_{\tau}(s_1,a_1,\eta_2))]
    \end{align*}
    \normalsize
    Based on the definition of $Q_{\tau}^{\pi_2}(s_1,a_1,\eta_2)$, we have
    \begin{align*}
        \nabla_{\theta_2}J^{\pi}_{\tau}(s_1)
       = &\nabla_{\theta_2}(\sum_{a_1,\eta_2}\pi_1(a_1,\eta_2|s_1)(\tau\log\pi_1(a_1,\eta_2|s_1)+Q^{\pi}_{\tau}(s_1,a_1,\eta_2)))\\
    =&\sum_{a_1,\eta_2}(\pi_1(a_1,\eta_2|s_1)\nabla_{\theta_2}(\bar{C}_1(s_1,a_1,\eta_2)+\gamma\mathbb{E}_{s_2}[\widehat{J}^{\pi_2}_{\tau}(s_2,\eta_2)]))\\
    =& \gamma\sum_{a_1,\eta_2}\pi_1(a_1,\eta_2|s_1)\sum_{s_2}P(s_2|s_1,a_1)\nabla_{\theta_2}\widehat{J}^{\pi_2}_{\tau}(s_2,\eta_2)
    = \gamma\sum_{s_2,\eta_2}\text{Pr}^{\pi_1}(s_2,\eta_2|s_1)\nabla_{\theta_2}\widehat{J}^{\pi_2}_{\tau}(s_2,\eta_2)
    \end{align*}
    Now for $\nabla_{\theta_2}\widehat{J}^{\pi_2}_{\tau}(s_2,\eta_2)$, we have
    \begin{align*}
            &\nabla_{\theta_2}\widehat{J}^{\pi_2}_{\tau}(s_2,\eta_2)
            =\nabla_{\theta_2}\left(\sum_{a_2,\eta_3}\pi_2(a_2,\eta_3|s_2,\eta_2)(\tau\log\pi_2(a_2,\eta_3|s_2,\eta_2)+\widehat{Q}_{\tau}^{\pi_2}(s_2,\eta_2,a_2,\eta_3))\right)\\
    =&\sum_{a_2,\eta_3}\Big(\nabla_{\theta_2}\pi_2(a_2,\eta_3|s_2,\eta_2)(\tau\log\pi_2(a_2,\eta_3|s_2,\eta_2)+\widehat{Q}_{\tau}^{\pi_2}(s_2,\eta_2,a_2,\eta_3))\\ &+\pi_2(a_2,\eta_3|s_2,\eta_2)\nabla_{\theta_2}(\tau\log\pi_2(a_2,\eta_3|s_2,\eta_2)+\bar{C}(s_2,\eta_2,a_2,\eta_{3})+\gamma\mathbb{E}_{s_{3}\sim P(\cdot|s_2,a_2)}[\widehat{J}^{\pi_2}_{\tau}(s_{3},\eta_{3})])\Big)\\
    \overset{(a)}{=}&\sum_{a_2,\eta_3}(\nabla_{\theta_2}\pi_2(a_2,\eta_3|s_2,\eta_2)(\tau\log\pi_2(a_2,\eta_3|s_2,\eta_2)+\widehat{Q}_{\tau}^{\pi_2}(s_2,\eta_2,a_2,\eta_3)))\\
    &+ \gamma\sum_{a_2,\eta_3}\pi_2(a_2,\eta_3|s_2,\eta_2)\sum_{s_3}P(s_3|s_2,a_2)\nabla_{\theta_2}\widehat{J}_{\tau}^{\pi_2}(s_3,\eta_3),
    \end{align*}
    where $(a)$ is true because $\sum_{a_2,\eta_3}\pi_2(a_2,\eta_3|s_2,\eta_2)\nabla_{\theta_2}(\tau\log\pi_2(a_2,\eta_3|s_2,\eta_2))=0$.
    Using a similar argument in risk-neutral PG theorems \citep{williams1992simple,sutton1999policy}, we obtain
        \begin{align*}
        &\nabla_{\theta_2}\widehat{J}^{\pi_2}_{\tau}(s_2,\eta_2)\\
        =&\frac{1}{1-\gamma}\mathbb{E}_{(s_t,\eta_t)\sim d^{\pi}_{s_2,\eta_2}}\mathbb{E}_{(a_t,\eta_{t+1})\sim\pi_2(\cdot|s_t,\eta_t)}[\nabla_{\theta_2}\log\pi_2(a_t,\eta_{t+1}|s_t,\eta_t)(\tau\log\pi_2(a_t,\eta_{t+1}|s_t,\eta_t)+\widehat{Q}_{\tau}^{\pi_2})]\\
        \overset{(a)}{=}&\frac{1}{1-\gamma}\mathbb{E}_{(s_t,\eta_t)\sim d^{\pi}_{s_2,\eta_2}}\mathbb{E}_{(a_t,\eta_{t+1})\sim\pi_2(\cdot|s_t,\eta_t)}[\nabla_{\theta_2}\log\pi_2(a_t,\eta_{t+1}|s_t,\eta_t)(-\widehat{A}^{\pi_2}_{\tau}(s_t,\eta_t,a_t,\eta_{t+1}))]
    \end{align*}
    \normalsize
    where $(a)$ is based on Eq.~\eqref{eq:natural-advantage2} and $\mathbb{E}_{(a_t,\eta_{t+1})\sim\pi_2(\cdot|s_t,\eta_t)}[\nabla_{\theta_2}\log\pi_2(a_t,\eta_{t+1}|s_t,\eta_t)(-\widehat{J}^{\pi}_{\tau}(s_t,\eta_t))]=0$.
    As a result,
        \begin{align*}
    &\nabla_{\theta_2}J^{\pi}_{\tau}(\rho)
=\gamma\sum_{s_2,\eta_2}\sum_{s_1}\rho(s_1)\text{Pr}^{\pi_1}(s_2,\eta_2|s_1)\nabla_{\theta_2}\widehat{J}^{\pi_2}_{\tau}(s_2,\eta_2)\\
    =&\gamma\sum_{s_2,\eta_2}\rho^{\pi}(s_2,\eta_2)\nabla_{\theta_2}\widehat{J}^{\pi_2}_{\tau}(s_2,\eta_2)\\
    =&\frac{\gamma}{1-\gamma}\mathbb{E}_{(s_t,\eta_t)\sim d^{\pi}_{\rho^{\pi}}}\mathbb{E}_{(a_t,\eta_{t+1})\sim\pi_2(\cdot|s_t,\eta_t)}[\nabla_{\theta_2}\log\pi_2(a_t,\eta_{t+1}|s_t,\eta_t)(-\widehat{A}^{\pi_2}_{\tau}(s_t,\eta_t,a_t,\eta_{t+1}))].
    \end{align*}
    This completes the proof.
  \endproof

\proof{\textbf{of Lemma \ref{lem:pseudoinverse}}}
    Following Appendix C.6 in \cite{cen2022fast} and using the definition of Moore-Penrose pseudoinverse, we know $[(\mathcal{F}^{\theta_1}_{\rho})^{\dagger}\nabla_{\theta_1}J_{\tau}^{\pi}(\rho)]$ is the optimal solution to the following least-square problem
$\min_{w\in\mathbb{R}^{|\mathcal{S}||\mathcal{A}||\mathcal{H}|}}||\mathcal{F}_{\rho}^{\theta_1}w-\nabla_{\theta_1}J_{\tau}^{\pi}(\rho)||_2^2$.
    Now from Eq.~\eqref{eq:fisher1}, we have
$\mathcal{F}_{\rho}^{\theta_1}w=\mathbb{E}_{s\sim\rho}\mathbb{E}_{a,\eta\sim\pi_1}[(\nabla_{\theta_1}\log\pi_1(a,\eta|s))(\nabla_{\theta_1}\log\pi_1(a,\eta|s))^{\mathsf T}w]$
    for any fixed vector $w=[w_{s,a,\eta}]_{(s,a,\eta)\in\mathcal{S}\times\mathcal{A}\times\mathcal{H}}$. As a result, for any $(s,a,\eta)\in\mathcal{S}\times\mathcal{A}\times\mathcal{H}$, one has
    \small
    \begin{align*}
&(\mathcal{F}_{\rho}^{\theta_1}w)_{s,a,\eta}
={\color{black}\kappa_1}\mathbb{E}_{s^{\prime}\sim\rho}\mathbb{E}_{a^{\prime},\eta^{\prime}\sim\pi_1(\cdot|s^{\prime})}\left[\frac{\partial\log\pi_1(a^{\prime},\eta^{\prime}|s^{\prime})}{\partial\theta_1(s,a,\eta)}\left(\sum_{\widetilde{s},\widetilde{a},\widetilde{\eta}}\frac{\partial\log\pi_1(a^{\prime},\eta^{\prime}|s^{\prime})}{\partial\theta_1(\widetilde{s},\widetilde{a},\widetilde{\eta})}w_{\widetilde{s},\widetilde{a},\widetilde{\eta}}\right)\right]\\
=&{\color{black}\kappa_1}\mathbb{E}_{s^{\prime}\sim\rho}\mathbb{E}_{a^{\prime},\eta^{\prime}\sim\pi_1(\cdot|s^{\prime})}\left[\textbf{1}(s^{\prime}=s)(\textbf{1}(a^{\prime}=a,\eta^{\prime}=\eta)-\pi_1(a,\eta|s))\left(\sum_{\widetilde{s},\widetilde{a},\widetilde{\eta}}\textbf{1}(\widetilde{s}=s^{\prime})\left(\textbf{1}(\widetilde{a}=a',\widetilde{\eta}=\eta')-\pi_1(\widetilde{a},\widetilde{\eta}|\widetilde{s})\right)w_{\widetilde{s},\widetilde{a},\widetilde{\eta}}\right)\right]\\
=&{\color{black}\kappa_1}\mathbb{E}_{s^{\prime}\sim\rho}\mathbb{E}_{a^{\prime},\eta^{\prime}\sim\pi_1(\cdot|s^{\prime})}\left[\textbf{1}(s^{\prime}=s)(\textbf{1}(a^{\prime}=a,\eta^{\prime}=\eta)-\pi_1(a,\eta|s))\left(w_{s^{\prime},a^{\prime},\eta^{\prime}}-\sum_{\widetilde{a},\widetilde{\eta}}\pi_1(\widetilde{a},\widetilde{\eta}|s^{\prime})w_{s^{\prime},\widetilde{a},\widetilde{\eta}}\right)\right]\\
=&{\color{black}\kappa_1}\rho(s)\mathbb{E}_{a^{\prime},\eta^{\prime}\sim\pi_1(\cdot|s)}\left[(\textbf{1}(a^{\prime}=a,\eta^{\prime}=\eta)-\pi_1(a,\eta|s))\left(w_{s,a^{\prime},\eta^{\prime}}-c(s)\right)\right]\\
=&{\color{black}\kappa_1}\rho(s)\mathbb{E}_{a^{\prime},\eta^{\prime}\sim\pi_1(\cdot|s)}\left[\textbf{1}(a^{\prime}=a,\eta^{\prime}=\eta)w_{s,a^{\prime},\eta^{\prime}}-\pi_1(a,\eta|s)w_{s,a^{\prime},\eta^{\prime}}-\textbf{1}(a^{\prime}=a,\eta^{\prime}=\eta)c(s)+\pi_1(a,\eta|s)c(s)\right]\\
=&{\color{black}\kappa_1}\rho(s)\left[\pi_1(a,\eta|s)w_{s,a,\eta}-\pi_1(a,\eta|s)c(s)-\pi_1(a,\eta|s)c(s)+\pi_1(a,\eta|s)c(s)\right]\\
=&{\color{black}\kappa_1}\rho(s)\pi_1(a,\eta|s)(w_{s,a,\eta}-c(s))
    \end{align*}
    \normalsize
    where we define $c(s)=\sum_{{a},{\eta}}\pi_1({a},{\eta}|s)w_{s,{a},{\eta}}$. Using Theorem \ref{thm:natural-gradient}, we have 
    \begin{align*}
       & \frac{\partial J_{\tau}^{\pi}(\rho)}{\partial \theta_1(s,a,\eta)}=\rho(s)\pi_1(a,\eta|s)(-A_{\tau}^{\pi}(s,a,\eta))\\
        & \frac{\partial J_{\tau}^{\pi}(\rho)}{\partial \theta_2(s,\eta,a,\eta^{\prime})}=\frac{\gamma}{1-\gamma}d_{\rho^{\pi}}^{\pi}(s,\eta)\pi_2(a,\eta^{\prime}|s,\eta)(-\widehat{A}_{\tau}^{\pi}(s,\eta,a,\eta^{\prime}))
    \end{align*}
    
    Consequently, we have
    \begin{align*}
        ||\mathcal{F}_{\rho}^{\theta_1}w-\nabla_{\theta_1}J_{\tau}^{\pi}(\rho)||_2^2&=\sum_{s,a,\eta}\left({\color{black}\kappa_1}\rho(s)\pi_1(a,\eta|s)(w_{s,a,\eta}-c(s))-\rho(s)\pi_1(a,\eta|s)(-A_{\tau}^{\pi}(s,a,\eta))\right)^2\\
        &=\sum_{s,a,\eta}\left({\color{black}\kappa_1}\rho(s)\pi_1(a,\eta|s)(w_{s,a,\eta}-c(s)+{\color{black}\frac{1}{\kappa_1}}A_{\tau}^{\pi}(s,a,\eta))\right)^2
    \end{align*}
    which is minimized by choosing $w_{s,a,\eta}=-{\color{black}\frac{1}{\kappa_1}}A_{\tau}^{\pi}(s,a,\eta)+c(s)$. Thus, we have $[(\mathcal{F}^{\theta_1}_{\rho})^{\dagger}\nabla_{\theta_1}J_{\tau}^{\pi}(\rho)](s,a,\eta)=-{\color{black}\frac{1}{\kappa_1}}A_{\tau}^{\pi}(s,a,\eta)+c(s)$.

    Similarly, note that $[(\mathcal{F}^{\theta_2}_{\rho})^{\dagger}\nabla_{\theta_2}J_{\tau}^{\pi}(\rho)](s_t,\eta_t,a_t,\eta_{t+1})$ is the optimal solution to the following least-square problem
$\min_{w\in\mathbb{R}^{|\mathcal{S}||\mathcal{A}||\mathcal{H}|^2}}||\mathcal{F}_{\rho}^{\theta_2}w-\nabla_{\theta_2}J_{\tau}^{\pi}(\rho)||_2^2$.
    Following the same logic, one can show that 
$(\mathcal{F}_{\rho}^{\theta_2}w)_{s_t,\eta_t,a_t,\eta_{t+1}}={\color{black}\kappa_2}d_{\rho^{\pi}}^{\pi}(s_t,\eta_t)\pi_2(a_t,\eta_{t+1}|s_t,a_t)(w_{s_t,\eta_t,a_t,\eta_{t+1}}-c(s_t,\eta_t))$,
    where $c(s_t,\eta_t)=\sum_{a_t,\eta_{t+1}}\pi_2(a_t,\eta_{t+1}|s_t,\eta_t)w_{s_t,\eta_t,a_t,\eta_{t+1}}$. As a result, we have
    \begin{align*}
        &||\mathcal{F}_{\rho}^{\theta_2}w-\nabla_{\theta_2}J_{\tau}^{\pi}(\rho)||_2^2\\
        =&\sum_{s_t,\eta_t,a_t,\eta_{t+1}}\left({\color{black}\kappa_2}d_{\rho^{\pi}}^{\pi}(s_t,\eta_t)\pi_2(a_t,\eta_{t+1}|s_t,\eta_t)(w_{s_t,\eta_t,a_t,\eta_{t+1}}-c(s_t,\eta_t)+\frac{\gamma}{\kappa_2(1-\gamma)}\widehat{A}_{\tau}^{\pi}(s_t,\eta_t,a_t,\eta_{t+1}))\right)^2
    \end{align*}
    which is minimized by setting $w_{s_t,\eta_t,a_t,\eta_{t+1}}=-\frac{\gamma}{\kappa_2(1-\gamma)}\widehat{A}_{\tau}^{\pi}(s_t,\eta_t,a_t,\eta_{t+1})+c(s_t,\eta_t)$. 
    This completes the proof.
  \endproof

\proof{\textbf{of Lemma \ref{lem:NPG-update}}}
    Based on the softmax parameterization, we have
    \begin{align*}
        \pi_1^{(t+1)}(a_1,\eta_2|s_1)\propto&\exp{(\theta_1^{(t+1)}(s_1,a_1,\eta_2))}=\exp{\left(\theta_1^{(t)}(s_1,a_1,\eta_2)- {\color{black}\beta}[(\mathcal{F}^{\theta_1^{(t)}}_{\rho})^{\dagger}\nabla_{\theta_1}J_{\tau}^{\pi}(\rho)](s_1,a_1,\eta_2)\right)}\\
        \overset{(a)}{\propto}&\pi_1^{(t)}(a_1,\eta_2|s_1)\exp{\left(\frac{{\color{black}\beta}}{\kappa_1}(A_{\tau}^{(t)}(s_1,a_1,\eta_2)-c(s_1))\right)}\\
        \overset{(b)}{\propto}& \pi_1^{(t)}(a_1,\eta_2|s_1)\exp{\left(\frac{\beta}{\kappa_1}(J_{\tau}^{(t)}(s_1)-\tau\log\pi_1^{(t)}(a_1,\eta_2|s_1)-Q_{\tau}^{(t)}(s_1,a_1,\eta_2))\right)}\\
        \overset{(c)}{\propto}& \left(\pi_1^{(t)}(a_1,\eta_2|s_1)\right)^{1-\frac{\beta\tau}{\kappa_1}}\exp{\left(-\frac{\beta}{\kappa_1}Q_{\tau}^{(t)}(s_1,a_1,\eta_2)\right)}
    \end{align*}
    where $(a)$ uses Eq.~\eqref{eq:pseudo1}, and $(b)$ and $(c)$ use the fact that $c(s_1)$ and $J_{\tau}^{(t)}(s_1)$ do not depend on $a_1,\eta_2$.

    Similarly, we have
    \begin{align*}
        &\pi_2^{(t+1)}(a_i,\eta_{i+1}|s_i,\eta_i)
        \propto\exp{(\theta_2^{(t+1)}(a_i,\eta_{i+1}|s_i,\eta_i))}\\
        =&\exp{\left(\theta_2^{(t)}(s_i,\eta_i,a_i,\eta_{i+1}) - \beta[(\mathcal{F}^{\theta_2^{(t)}}_{\rho})^{\dagger}\nabla_{\theta_2}J_{\tau}^{\pi}(\rho)](s_i,\eta_i,a_i,\eta_{i+1})\right)}\\
        \propto&\pi_2^{(t)}(a_i,\eta_{i+1}|s_i,\eta_i)\exp{\left(\frac{\beta\gamma}{\kappa_2(1-\gamma)}\widehat{A}^{(t)}_{\tau}(s_i,\eta_i,a_i,\eta_{i+1})-\beta c(s_i,\eta_i)\right)}\\
        \propto& \pi_2^{(t)}(a_i,\eta_{i+1}|s_i,\eta_i)\exp{\left(\frac{\beta\gamma}{\kappa_2(1-\gamma)}(\widehat{J}^{(t)}_{\tau}(s_i,\eta_i)-\tau\log\pi_2^{(t)}(a_i,\eta_{i+1}|s_i,\eta_i)-\widehat{Q}_{\tau}^{(t)}(s_i,\eta_i,a_i,\eta_{i+1}))\right)}\\
        \propto& \left(\pi_2^{(t)}(a_i,\eta_{i+1}|s_i,\eta_i)\right)^{1-\frac{\beta\tau\gamma}{\kappa_2(1-\gamma)}}\exp{\left(-\frac{\beta\gamma}{\kappa_2(1-\gamma)}\widehat{Q}_{\tau}^{(t)}(s_i,\eta_i,a_i,\eta_{i+1})\right)}
    \end{align*}
    This completes the proof.
  \endproof

\proof{\textbf{of Theorem \ref{thm:performance-improvement-NPG}}}
    We first prove the performance improvement for value function $\widehat{J}_{\tau}^{(t)}(s_2,\eta_2)$ (Eq.~\eqref{eq:improvement2}). From Eq.~\eqref{eq:NPG-update2}, we have 
    \begin{align*}
        \log\pi_2^{(t+1)}(a_i,\eta_{i+1}|s_i,\eta_i)=(1-{\color{black}\frac{\beta\tau\gamma}{\kappa_2(1-\gamma)}})\log\pi^{(t)}_2(a_i,\eta_{i+1}|s_i,\eta_i)-{\color{black}\frac{\beta\gamma}{\kappa_2(1-\gamma)}}\widehat{Q}_{\tau}^{(t)}(s_i,\eta_i,a_i,\eta_{i+1})-\log Z^{(t)}_2(s_i,\eta_i)
    \end{align*}
    Rearranging the terms gives us {\color{black}(when $\frac{\gamma}{\kappa_2}\not=0$)}
    \begin{align}
        &\tau\log\pi_2^{(t)}(a_i,\eta_{i+1}|s_i,\eta_i)+\widehat{Q}_{\tau}^{(t)}(s_i,\eta_i,a_i,\eta_{i+1})\nonumber\\
        =&-{\color{black}\frac{\kappa_2(1-\gamma)}{\beta\gamma}}\log Z^{(t)}_2(s_i,\eta_i)-{\color{black}\frac{\kappa_2(1-\gamma)}{\beta\gamma}}\left(\log\pi_2^{(t+1)}(a_i,\eta_{i+1}|s_i,\eta_i)-\log\pi^{(t)}_2(a_i,\eta_{i+1}|s_i,\eta_i)\right)\label{eq:normalization-Z}
    \end{align}
    As a result, we have
    \begin{align}
        &\widehat{J}_{\tau}^{(t)}(s_2,\eta_2)
        =\mathbb{E}_{a_2,\eta_3\sim\pi_2^{(t)}(\cdot|s_2,\eta_2)}\left[\tau\log\pi_2^{(t)}(a_2,\eta_3|s_2,\eta_2)+\widehat{Q}_{\tau}^{(t)}(s_2,\eta_2,a_2,\eta_3)\right]\nonumber\\
        =&\mathbb{E}_{a_2,\eta_3\sim\pi_2^{(t)}(\cdot|s_2,\eta_2)}\left[-{\color{black}\frac{\kappa_2(1-\gamma)}{\beta\gamma}}\log Z^{(t)}_2(s_2,\eta_2)\right]\nonumber\\
        &+\mathbb{E}_{a_2,\eta_3\sim\pi_2^{(t)}(\cdot|s_2,\eta_2)}\left[-{\color{black}\frac{\kappa_2(1-\gamma)}{\beta\gamma}}(\log\pi_2^{(t+1)}(a_2,\eta_{3}|s_2,\eta_2)-\log\pi^{(t)}_2(a_2,\eta_{3}|s_2,\eta_2))\right]\nonumber\\
        =&\mathbb{E}_{a_2,\eta_3\sim\pi_2^{(t+1)}(\cdot|s_2,\eta_2)}\left[-{\color{black}\frac{\kappa_2(1-\gamma)}{\beta\gamma}}\log Z^{(t)}_2(s_2,\eta_2)\right]+{\color{black}\frac{\kappa_2(1-\gamma)}{\beta\gamma}}\text{KL}(\pi_2^{(t)}(\cdot|s_2,\eta_2)||\pi_2^{(t+1)}(\cdot|s_2,\eta_2))\label{eq:NPG-performance-KL}\\
        \overset{(a)}{=}& \mathbb{E}_{a_2,\eta_3\sim\pi_2^{(t+1)}(\cdot|s_2,\eta_2)}\left[\tau\log\pi_2^{(t+1)}(a_2,\eta_3|s_2,\eta_2)+\widehat{Q}_{\tau}^{(t)}(s_2,\eta_2,a_2,\eta_3)+(-\tau+{\color{black}\frac{\kappa_2(1-\gamma)}{\beta\gamma}})(\log\pi_2^{(t+1)}-\log\pi_2^{(t)})\right]\nonumber\\
        &+{\color{black}\frac{\kappa_2(1-\gamma)}{\beta\gamma}}\text{KL}(\pi_2^{(t)}(\cdot|s_2,\eta_2)||\pi_2^{(t+1)}(\cdot|s_2,\eta_2))\nonumber\\
        =& \mathbb{E}_{\substack{a_2,\eta_3\sim\pi_2^{(t+1)}(\cdot|s_2,\eta_2)\\s_3\sim P(\cdot|s_2,a_2)}}\left[\tau\log\pi_2^{(t+1)}(a_2,\eta_3|s_2,\eta_2)+\bar{C}(s_2,\eta_2,a_2,\eta_3)+\gamma\widehat{J}_{\tau}^{(t)}(s_3,\eta_3)\right]\nonumber\\
        &+(-\tau+{\color{black}\frac{\kappa_2(1-\gamma)}{\beta\gamma}})\text{KL}(\pi_2^{(t+1)}(\cdot|s_2,\eta_2)||\pi_2^{(t)}(\cdot|s_2,\eta_2))+{\color{black}\frac{\kappa_2(1-\gamma)}{\beta\gamma}}\text{KL}(\pi_2^{(t)}(\cdot|s_2,\eta_2)||\pi_2^{(t+1)}(\cdot|s_2,\eta_2))\nonumber\\
        =& \mathbb{E}_{\substack{a_i,\eta_{i+1}\sim\pi_2^{(t+1)}(\cdot|s_i,\eta_i)\\s_{i+1}\sim P(\cdot|s_i,a_i),\ i\ge 2}}\Big[\sum_{i=2}^{\infty}\gamma^{i-2}\{\tau\log\pi_2^{(t+1)}(a_i,\eta_{i+1}|s_i,\eta_i)+\bar{C}(s_i,\eta_i,a_i,\eta_{i+1})\}\nonumber\\
        &+\sum_{i=2}^{\infty}\gamma^{i-2}\{(-\tau+{\color{black}\frac{\kappa_2(1-\gamma)}{\beta\gamma}})\text{KL}(\pi_2^{(t+1)}(\cdot|s_i,\eta_i)||\pi_2^{(t)}(\cdot|s_i,\eta_i))+{\color{black}\frac{\kappa_2(1-\gamma)}{\beta\gamma}}\text{KL}(\pi_2^{(t)}(\cdot|s_i,\eta_i)||\pi_2^{(t+1)}(\cdot|s_i,\eta_i))\}\Big]\label{eq:NPG-performance-recursive}\\
        \overset{(b)}{=}&\widehat{J}^{(t+1)}_{\tau}(s_2,\eta_2)+\frac{1}{1-\gamma}\mathbb{E}_{(s_i,\eta_i)\sim d_{(s_2,\eta_2)}^{{(t+1)}}}\Big[(-\tau+{\color{black}\frac{\kappa_2(1-\gamma)}{\beta\gamma}})\text{KL}(\pi_2^{(t+1)}(\cdot|s_i,\eta_i)||\pi_2^{(t)}(\cdot|s_i,\eta_i))\nonumber\\
        &+{\color{black}\frac{\kappa_2(1-\gamma)}{\beta\gamma}}\text{KL}(\pi_2^{(t)}(\cdot|s_i,\eta_i)||\pi_2^{(t+1)}(\cdot|s_i,\eta_i))\Big]\nonumber\\
        {=}&\widehat{J}^{(t+1)}_{\tau}(s_2,\eta_2)+\mathbb{E}_{(s_i,\eta_i)\sim d_{(s_2,\eta_2)}^{{(t+1)}}}\Big[(-\frac{\tau}{1-\gamma}+\frac{\kappa_2}{\beta\gamma})\text{KL}(\pi_2^{(t+1)}(\cdot|s_i,\eta_i)||\pi_2^{(t)}(\cdot|s_i,\eta_i))\nonumber\\
        &+\frac{\kappa_2}{\beta\gamma}\text{KL}(\pi_2^{(t)}(\cdot|s_i,\eta_i)||\pi_2^{(t+1)}(\cdot|s_i,\eta_i))\Big]\label{eq:performance-improve1}
    \end{align}
    where $(a)$ is due to Eq.~\eqref{eq:normalization-Z} and $(b)$ is true because $\widehat{J}^{(t+1)}_{\tau}(s_2,\eta_2)$ can be viewed as the value function of $\pi^{(t+1)}$ with regularized cost $\tau\log\pi_2^{(t+1)}(a_i,\eta_{i+1}|s_i,\eta_i)+\bar{C}(s_i,\eta_i,a_i,\eta_{i+1})$.
    Next, we prove the performance improvement of value function $J^{(t)}_{\tau}(s_1)$ (Eq.~\eqref{eq:improvement1}). From Eq.~\eqref{eq:NPG-update1}, we have
    \begin{align*}
        \log\pi_1^{(t+1)}(a_1,\eta_2|s_1)=(1-\frac{\beta\tau}{\kappa_1})\log\pi^{(t)}_1(a_1,\eta_2|s_1)-\frac{\beta}{\kappa_1}{Q}_{\tau}^{(t)}(s_1,a_1,\eta_2)-\log Z^{(t)}_1(s_1)
    \end{align*}
    Rearranging the terms gives us
    \begin{align*}
        &\tau\log\pi_1^{(t)}(a_1,\eta_2|s_1)+{Q}_{\tau}^{(t)}(s_1,a_1,\eta_2)
        =-{\color{black}\frac{\kappa_1}{\beta}}\log Z^{(t)}_1(s_1)-{\color{black}\frac{\kappa_1}{\beta}}(\log\pi_1^{(t+1)}(a_1,\eta_2|s_1)-\log\pi^{(t)}_1(a_1,\eta_2|s_1))
    \end{align*}
    As a result, we have
    \begin{align}
                &{J}_{\tau}^{(t)}(s_1)
        =\mathbb{E}_{a_1,\eta_2\sim\pi_1^{(t)}(\cdot|s_1)}\left[\tau\log\pi_1^{(t)}(a_1,\eta_2|s_1)+{Q}_{\tau}^{(t)}(s_1,a_1,\eta_2)\right]\nonumber\\
        =&\mathbb{E}_{a_1,\eta_2\sim\pi_1^{(t)}(\cdot|s_1)}\left[-{\color{black}\frac{\kappa_1}{\beta}}\log Z^{(t)}_1(s_1)\right]+\mathbb{E}_{a_1,\eta_2\sim\pi_1^{(t)}(\cdot|s_1)}\left[-{\color{black}\frac{\kappa_1}{\beta}}(\log\pi_1^{(t+1)}(a_1,\eta_{2}|s_1)-\log\pi^{(t)}_1(a_1,\eta_{2}|s_1))\right]\nonumber\\
        =&\mathbb{E}_{a_1,\eta_2\sim\pi_1^{(t+1)}(\cdot|s_1)}\left[-{\color{black}\frac{\kappa_1}{\beta}}\log Z^{(t)}_1(s_1)\right]+{\color{black}\frac{\kappa_1}{\beta}}\text{KL}(\pi_1^{(t)}(\cdot|s_1)||\pi_1^{(t+1)}(\cdot|s_1))\label{eq:J_tau_t}\\
        =& \mathbb{E}_{a_1,\eta_2\sim\pi_1^{(t+1)}(\cdot|s_1)}\left[\tau\log\pi_1^{(t+1)}(a_1,\eta_2|s_1)+{Q}_{\tau}^{(t)}(s_1,a_1,\eta_2)+(-\tau+{\color{black}\frac{\kappa_1}{\beta}})(\log\pi_1^{(t+1)}-\log\pi_1^{(t)})\right]\nonumber\\
        &+{\color{black}\frac{\kappa_1}{\beta}}\text{KL}(\pi_1^{(t)}(\cdot|s_1)||\pi_1^{(t+1)}(\cdot|s_1))\nonumber\\
        =& \mathbb{E}_{\substack{a_1,\eta_2\sim\pi_1^{(t+1)}(\cdot|s_1)\\s_2\sim P(\cdot|s_1,a_1)}}\left[\tau\log\pi_1^{(t+1)}(a_1,\eta_2|s_1)+\bar{C}_1(s_1,a_1,\eta_2)+\gamma\widehat{J}_{\tau}^{(t)}(s_2,\eta_2)\right]\nonumber\\
        &+(-\tau+{\color{black}\frac{\kappa_1}{\beta}})\text{KL}(\pi_1^{(t+1)}(\cdot|s_1)||\pi_1^{(t)}(\cdot|s_1))+{\color{black}\frac{\kappa_1}{\beta}}\text{KL}(\pi_1^{(t)}(\cdot|s_1)||\pi_1^{(t+1)}(\cdot|s_1))\nonumber\\
        \overset{(a)}{=}&\mathbb{E}_{\substack{a_1,\eta_2\sim\pi_1^{(t+1)}(\cdot|s_1)\\s_2\sim P(\cdot|s_1,a_1)}}\left[\tau\log\pi_1^{(t+1)}(a_1,\eta_2|s_1)+\bar{C}_1(s_1,a_1,\eta_2)+\gamma\widehat{J}^{(t+1)}_{\tau}(s_2,\eta_2)\right]\nonumber\\
        &+\mathbb{E}_{\substack{a_1,\eta_2\sim\pi_1^{(t+1)}(\cdot|s_1)\\s_2\sim P(\cdot|s_1,a_1)\\(s_i,\eta_i)\sim d_{(s_2,\eta_2)}^{{(t+1)}}}}\left[(-\frac{\tau\gamma}{1-\gamma}+\frac{\kappa_2}{\beta})\text{KL}(\pi_2^{(t+1)}(\cdot|s_i,\eta_i)||\pi_2^{(t)}(\cdot|s_i,\eta_i))+\frac{\kappa_2}{\beta}\text{KL}(\pi_2^{(t)}(\cdot|s_i,\eta_i)||\pi_2^{(t+1)}(\cdot|s_i,\eta_i))\right]\nonumber\\
        &+(-\tau+{\color{black}\frac{\kappa_1}{\beta}})\text{KL}(\pi_1^{(t+1)}(\cdot|s_1)||\pi_1^{(t)}(\cdot|s_1))+{\color{black}\frac{\kappa_1}{\beta}}\text{KL}(\pi_1^{(t)}(\cdot|s_1)||\pi_1^{(t+1)}(\cdot|s_1))\nonumber\\
        =&{J}_{\tau}^{(t+1)}(s_1)\nonumber\\
&+\mathbb{E}_{\substack{a_1,\eta_2\sim\pi_1^{(t+1)}(\cdot|s_1)\\s_2\sim P(\cdot|s_1,a_1)\\(s_i,\eta_i)\sim d_{(s_2,\eta_2)}^{{(t+1)}}}}\left[(-\frac{\tau\gamma}{1-\gamma}+\frac{\kappa_2}{\beta})\text{KL}(\pi_2^{(t+1)}(\cdot|s_i,\eta_i)||\pi_2^{(t)}(\cdot|s_i,\eta_i))+\frac{\kappa_2}{\beta}\text{KL}(\pi_2^{(t)}(\cdot|s_i,\eta_i)||\pi_2^{(t+1)}(\cdot|s_i,\eta_i))\right]\nonumber\\
        &+(-\tau+{\color{black}\frac{\kappa_1}{\beta}})\text{KL}(\pi_1^{(t+1)}(\cdot|s_1)||\pi_1^{(t)}(\cdot|s_1))+{\color{black}\frac{\kappa_1}{\beta}}\text{KL}(\pi_1^{(t)}(\cdot|s_1)||\pi_1^{(t+1)}(\cdot|s_1))\nonumber
    \end{align}
    where $(a)$ is due to Eq.~\eqref{eq:performance-improve1}.
    This completes the proof.
  \endproof

\proof{\textbf{of Proposition \ref{prop:linear-system}}}
The proof mainly follows from Section 4.2.2 in \cite{cen2022fast} with some minor changes.  We start by noticing that $A$ is a rank-1 matrix and has the following nice property:
    \begin{align}
        A=\begin{pmatrix}
            \gamma\\1
        \end{pmatrix}\begin{pmatrix}
            1-\omega, \omega
        \end{pmatrix} \text{ and } A^t=(1-\frac{\beta\tau\gamma}{\kappa_2})^{t-1}A,\ \forall t\ge 1\label{eq:decomposition-A}
    \end{align}
    which is true because $(1-\omega)\gamma+\omega=1-\frac{\beta\tau\gamma}{\kappa_2}$. The rest of the proof follows from \cite{cen2022fast}.
  \endproof

\proof{\textbf{of Lemma \ref{lem:approximate-NPG-improvement}}}
We first prove the performance improvement for value function $\widehat{J}_{\tau}^{(t)}(s_2,\eta_2)$.
    Recall that in approximate NPG updates, the policies are updated using the approximate $Q$-function, i.e., for any $(s_i,\eta_i,a_i,\eta_{i+1})\in \mathcal{S}\times\mathcal{H}\times\mathcal{A}\times\mathcal{H}$,
    \begin{align*}
        \pi_2^{(t+1)}(a_i,\eta_{i+1}|s_i,\eta_i)=\frac{1}{\widetilde{Z}_2^{(t)}(s_i,\eta_i)}\left(\pi_2^{(t)}(a_i,\eta_{i+1}|s_i,\eta_i)\right)^{1-{\color{black}\frac{\beta\tau\gamma}{\kappa_2(1-\gamma)}}}\exp{\left(-{\color{black}\frac{\beta\gamma}{\kappa_2(1-\gamma)}}\widetilde{\widehat{Q}}_{\tau}^{(t)}(s_i,\eta_i,a_i,\eta_{i+1})\right)}
    \end{align*}
    where $\widetilde{Z}_2^{(t)}(s_i,\eta_i)=\sum_{a_i,\eta_{i+1}}(\pi_2^{(t)}(a_i,\eta_{i+1}|s_i,\eta_i))^{1-{\color{black}\frac{\beta\tau\gamma}{\kappa_2(1-\gamma)}}}\exp{(-{\color{black}\frac{\beta\gamma}{\kappa_2(1-\gamma)}}\widetilde{\widehat{Q}}_{\tau}^{(t)}(s_i,\eta_i,a_i,\eta_{i+1}))}$. We also define an auxiliary policy sequence $\{\breve{\pi}_2^{(t)}\}$, which uses the exact soft $Q$-function of $\pi^{(t)}$ in the $t$-th iteration, i.e., for any $(s_i,\eta_i,a_i,\eta_{i+1})\in \mathcal{S}\times\mathcal{H}\times\mathcal{A}\times\mathcal{H}$,
    \begin{align}
        \breve{\pi}_2^{(t+1)}(a_i,\eta_{i+1}|s_i,\eta_i)=\frac{1}{{Z}_2^{(t)}(s_i,\eta_i)}\left({\pi}_2^{(t)}(a_i,\eta_{i+1}|s_i,\eta_i)\right)^{1-{\color{black}\frac{\beta\tau\gamma}{\kappa_2(1-\gamma)}}}\exp{\left(-{\color{black}\frac{\beta\gamma}{\kappa_2(1-\gamma)}}{\widehat{Q}}_{\tau}^{(t)}(s_i,\eta_i,a_i,\eta_{i+1})\right)}\label{eq:breve-pi}
    \end{align}
    where we abuse the notation by setting 
    $${Z}_2^{(t)}(s_i,\eta_i)=\sum_{a_i,\eta_{i+1}}\left({\pi}_2^{(t)}(a_i,\eta_{i+1}|s_i,\eta_i)\right)^{1-{\color{black}\frac{\beta\tau\gamma}{\kappa_2(1-\gamma)}}}\exp{\left(-{\color{black}\frac{\beta\gamma}{\kappa_2(1-\gamma)}}{\widehat{Q}}_{\tau}^{(t)}(s_i,\eta_i,a_i,\eta_{i+1})\right)}.$$
    Note that $\breve{\pi}_2^{(t+1)}$ is generated from ${\pi}_2^{(t)}$ instead of $\breve{\pi}_2^{(t)}$, since we assume that we only have one-step perfect update from a given policy ${\pi}_2^{(t)}$. We first observe that for any stepsize $0<\beta\le \min\{{\color{black}\frac{\kappa_2(1-\gamma)}{\tau\gamma},\frac{\kappa_1}{\tau}}\}$, we have
    \begin{align}
        &||\log\pi_2^{(t+1)}-\log\breve{\pi}_2^{(t+1)}||_{\infty}\nonumber\\
        \overset{(a)}{\le} & 2\left\|\log\left(\left(\pi_2^{(t)}\right)^{1-{\color{black}\frac{\beta\tau\gamma}{\kappa_2(1-\gamma)}}}\exp{\left(-{\color{black}\frac{\beta\gamma}{\kappa_2(1-\gamma)}}\widetilde{\widehat{Q}}_{\tau}^{(t)}\right)}\right)-\log\left(\left(\pi_2^{(t)}\right)^{1-{\color{black}\frac{\beta\tau\gamma}{\kappa_2(1-\gamma)}}}\exp{\left(-{\color{black}\frac{\beta\gamma}{\kappa_2(1-\gamma)}}{\widehat{Q}}_{\tau}^{(t)}\right)}\right)\right\|_{\infty}\nonumber\\
        \le & {\color{black}\frac{2\beta\gamma}{\kappa_2(1-\gamma)}}\left\|\widetilde{\widehat{Q}}_{\tau}^{(t)}-{\widehat{Q}}_{\tau}^{(t)}\right\|_{\infty}\label{eq:breve-pi_2}
    \end{align}
    where $(a)$ is due to Eq.~\eqref{eq:softmax-property}. From Eq.~\eqref{eq:breve-pi}, we also get
    \begin{align}
        &\tau\log\pi_2^{(t)}(a_i,\eta_{i+1}|s_i,\eta_i)+\widehat{Q}_{\tau}^{(t)}(s_i,\eta_i,a_i,\eta_{i+1})\nonumber\\
        =&-{\color{black}\frac{\kappa_2(1-\gamma)}{\beta\gamma}}\log Z^{(t)}_2(s_i,\eta_i)-{\color{black}\frac{\kappa_2(1-\gamma)}{\beta\gamma}}\left(\log\breve{\pi}_2^{(t+1)}(a_i,\eta_{i+1}|s_i,\eta_i)-\log\pi^{(t)}_2(a_i,\eta_{i+1}|s_i,\eta_i)\right)\label{eq:breve_pi_t+1}
    \end{align}
    Using the same reasoning as in Eq.~\eqref{eq:NPG-performance-KL}, we have
    \begin{align*}
        \widehat{J}_{\tau}^{(t)}(s_2,\eta_2)
        =&\mathbb{E}_{a_2,\eta_3\sim\pi_2^{(t)}(\cdot|s_2,\eta_2)}\left[\tau\log\pi_2^{(t)}(a_2,\eta_3|s_2,\eta_2)+\widehat{Q}_{\tau}^{(t)}(s_2,\eta_2,a_2,\eta_3)\right]\nonumber\\
        =&-\mathbb{E}_{a_2,\eta_3\sim\breve{\pi}_2^{(t+1)}(\cdot|s_2,\eta_2)}\left[{\color{black}\frac{\kappa_2(1-\gamma)}{\beta\gamma}}\log Z^{(t)}_2(s_2,\eta_2)\right]+{\color{black}\frac{\kappa_2(1-\gamma)}{\beta\gamma}}\text{KL}(\pi_2^{(t)}(\cdot|s_2,\eta_2)||\breve{\pi}_2^{(t+1)}(\cdot|s_2,\eta_2))\\
        =&-\mathbb{E}_{a_2,\eta_3\sim{\pi}_2^{(t+1)}(\cdot|s_2,\eta_2)}\left[{\color{black}\frac{\kappa_2(1-\gamma)}{\beta\gamma}}\log Z^{(t)}_2(s_2,\eta_2)\right]+{\color{black}\frac{\kappa_2(1-\gamma)}{\beta\gamma}}\text{KL}(\pi_2^{(t)}(\cdot|s_2,\eta_2)||\breve{\pi}_2^{(t+1)}(\cdot|s_2,\eta_2))
    \end{align*}
    The first term can be bounded as follows:
    \begin{align}
        &\mathbb{E}_{a_2,\eta_3\sim{\pi}_2^{(t+1)}(\cdot|s_2,\eta_2)}\left[{\color{black}\frac{\kappa_2(1-\gamma)}{\beta\gamma}}\log Z^{(t)}_2(s_2,\eta_2)\right]\nonumber\\
        \overset{(a)}{=}& \mathbb{E}_{a_2,\eta_3}\left[(\tau-{\color{black}\frac{\kappa_2(1-\gamma)}{\beta\gamma}})(\log\pi_2^{(t+1)}-\log\pi_2^{(t)})-{\color{black}\frac{\kappa_2(1-\gamma)}{\beta\gamma}}(\log\breve{\pi}_2^{(t+1)}-\log\pi_2^{(t+1)})-\tau\log\pi_2^{(t+1)}-\widehat{Q}_{\tau}^{(t)}\right]\nonumber\\
        \overset{(b)}{\le} & (\tau-{\color{black}\frac{\kappa_2(1-\gamma)}{\beta\gamma}})\text{KL}(\pi_2^{(t+1)}||\pi_2^{(t)})-\mathbb{E}_{a_2,\eta_3\sim{\pi}_2^{(t+1)}}\left[\tau\log\pi_2^{(t+1)}+\widehat{Q}_{\tau}^{(t)}\right]+2||\widetilde{\widehat{Q}}_{\tau}^{(t)}-{\widehat{Q}}_{\tau}^{(t)}||_{\infty}\label{eq:Z_2^t}
    \end{align}
    where $(a)$ is due to Eq.~\eqref{eq:breve_pi_t+1} and $(b)$ is due to Eq.~\eqref{eq:breve-pi_2}. As a result, when $0< \beta\le\min\{{\color{black}\frac{\kappa_2(1-\gamma)}{\tau\gamma},\frac{\kappa_1}{\tau}}\}$, we can bound $\widehat{J}_{\tau}^{(t)}(s_2,\eta_2)$ as follows:
    \begin{align}
        &\widehat{J}_{\tau}^{(t)}(s_2,\eta_2)\nonumber\\
        \ge & \mathbb{E}_{a_2,\eta_3}\left[\tau\log\pi_2^{(t+1)}+\widehat{Q}_{\tau}^{(t)}\right]-2||\widetilde{\widehat{Q}}_{\tau}^{(t)}-{\widehat{Q}}_{\tau}^{(t)}||_{\infty}+{\color{black}\frac{\kappa_2(1-\gamma)}{\beta\gamma}}\text{KL}(\pi_2^{(t)}||\breve{\pi}_2^{(t+1)})+({\color{black}\frac{\kappa_2(1-\gamma)}{\beta\gamma}}-\tau)\text{KL}(\pi_2^{(t+1)}||\pi_2^{(t)})\nonumber\\
        \ge &\mathbb{E}_{a_2,\eta_3}\left[\tau\log\pi_2^{(t+1)}+\widehat{Q}_{\tau}^{(t)}\right]-2||\widetilde{\widehat{Q}}_{\tau}^{(t)}-{\widehat{Q}}_{\tau}^{(t)}||_{\infty}\nonumber\\
    \overset{(a)}{=}&  \mathbb{E}_{a_2,\eta_3}\left[\tau\log\pi_2^{(t+1)}+\bar{C}(s_2,\eta_2,a_2,\eta_3)+\gamma\mathbb{E}_{s_3\sim P(\cdot|s_2,a_2)}[\widehat{J}_{\tau}^{(t)}(s_3,\eta_3)]\right]-2||\widetilde{\widehat{Q}}_{\tau}^{(t)}-{\widehat{Q}}_{\tau}^{(t)}||_{\infty}\nonumber\\
        \overset{(b)}{\ge} & \widehat{J}_{\tau}^{(t+1)}(s_2,\eta_2)-2\sum_{i=0}^{\infty}\gamma^i ||\widetilde{\widehat{Q}}_{\tau}^{(t)}-{\widehat{Q}}_{\tau}^{(t)}||_{\infty}\nonumber\\
        =&\widehat{J}_{\tau}^{(t+1)}(s_2,\eta_2)-\frac{2}{1-\gamma}||\widetilde{\widehat{{Q}}}_{\tau}^{(t)}-\widehat{Q}_{\tau}^{(t)}||_{\infty}\label{eq:hatJ-lowerbound}
    \end{align}
    where $(b)$ is by applying the inequality $(a)$ recursively as in Eq.~\eqref{eq:NPG-performance-recursive}. 

    Similarly, for any $(s,a,\eta)\in \mathcal{S}\times\mathcal{A}\times\mathcal{H}$, the approximate NPG updates 
        \begin{align*}
       \pi_1^{(t+1)}(a,\eta|s)=\frac{1}{\widetilde{Z}_1^{(t)}(s)}(\pi_1^{(t)}(a,\eta|s))^{1-{\color{black}\frac{\beta\tau}{\kappa_1}}}\exp\left(-{\color{black}\frac{\beta}{\kappa_1}}\widetilde{Q}_{\tau}^{(t)}(s,a,\eta)\right)
       \end{align*}
    We define an auxiliary policy sequence $\{\breve{\pi}_1^{(t)}\}$, which uses the exact soft $Q$-function of $\pi^{(t)}$ in the $t$-th iteration, i.e., for any $(s, a,\eta)\in \mathcal{S}\times\mathcal{A}\times\mathcal{H}$,
     \begin{align}
        \breve{\pi}_1^{(t+1)}(a,\eta|s)=\frac{1}{{Z}_1^{(t)}(s)}({\pi}_1^{(t)}(a,\eta|s))^{1-{\color{black}\frac{\beta\tau}{\kappa_1}}}\exp{\left(-{\color{black}\frac{\beta}{\kappa_1}}{{Q}}_{\tau}^{(t)}(s,a,\eta)\right)}\label{eq:breve-pi1}
    \end{align}
    where we abuse the notation by setting 
    $${Z}_1^{(t)}(s)=\sum_{a,\eta}({\pi}_1^{(t)}(a,\eta|s))^{1-{\color{black}\frac{\beta\tau}{\kappa_1}}}\exp{\left(-{\color{black}\frac{\beta}{\kappa_1}}{{Q}}_{\tau}^{(t)}(s,a,\eta)\right)}.$$ 
    Using the reasoning as in Eq.~\eqref{eq:breve-pi_2}, we get $||\log\pi_1^{(t+1)}-\log\breve{\pi}_1^{(t+1)}||_{\infty}\le {\color{black}\frac{2\beta}{\kappa_1}}||\widetilde{Q}_{\tau}^{(t)}-Q_{\tau}^{(t)}||_{\infty}$. Similarly, from Eq.~\eqref{eq:breve-pi1}, we have
     \begin{align}
        \tau\log\pi_1^{(t)}(a,\eta|s)+{Q}_{\tau}^{(t)}(s,a,\eta)=-\frac{\kappa_1}{\beta}\log Z^{(t)}_1(s)-\frac{\kappa_1}{\beta}\left(\log\breve{\pi}_1^{(t+1)}(a,\eta|s)-\log\pi^{(t)}_1(a,\eta|s)\right),\label{eq:breve_pi_1_t+1}
    \end{align}
    and following the reasoning in Eq.~\eqref{eq:J_tau_t}, we have
    \begin{align*}
        &{J}_{\tau}^{(t)}(s)\\
        =&\mathbb{E}_{a,\eta\sim\pi_1^{(t)}(\cdot|s)}\left[\tau\log\pi_1^{(t)}(a,\eta|s)+{Q}_{\tau}^{(t)}(s,a,\eta)\right]\nonumber\\
        \overset{(a)}{=}&-\mathbb{E}_{a,\eta\sim\breve{\pi}_1^{(t+1)}(\cdot|s)}\left[\frac{\kappa_1}{\beta}\log Z^{(t)}_1(s)\right]+\frac{\kappa_1}{\beta}\text{KL}(\pi_1^{(t)}(\cdot|s)||\breve{\pi}_1^{(t+1)}(\cdot|s))\\
        =&-\mathbb{E}_{a,\eta\sim{\pi}_1^{(t+1)}(\cdot|s)}\left[\frac{\kappa_1}{\beta}\log Z^{(t)}_1(s)\right]+\frac{\kappa_1}{\beta}\text{KL}(\pi_1^{(t)}(\cdot|s)||\breve{\pi}_1^{(t+1)}(\cdot|s))\\
        \overset{(b)}{\ge} &  \mathbb{E}_{a,\eta\sim{\pi}_1^{(t+1)}}\left[\tau\log\pi_1^{(t+1)}+{Q}_{\tau}^{(t)}\right]-2||\widetilde{{Q}}_{\tau}^{(t)}-{{Q}}_{\tau}^{(t)}||_{\infty}+\frac{\kappa_1}{\beta}\text{KL}(\pi_1^{(t)}||\breve{\pi}_1^{(t+1)})+\left(\frac{\kappa_1}{\beta}-\tau\right)\text{KL}(\pi_1^{(t+1)}||\pi_1^{(t)})\\
        \ge &\mathbb{E}_{a,\eta\sim{\pi}_1^{(t+1)}}\left[\tau\log\pi_1^{(t+1)}+{Q}_{\tau}^{(t)}\right]-2||\widetilde{{Q}}_{\tau}^{(t)}-{{Q}}_{\tau}^{(t)}||_{\infty}\\
        =&  \mathbb{E}_{a,\eta\sim{\pi}_1^{(t+1)}}\left[\tau\log\pi_1^{(t+1)}+\bar{C}(s,a,\eta)+\gamma\mathbb{E}_{s^{\prime}\sim P(\cdot|s,a)}[\widehat{J}_{\tau}^{(t)}(s^{\prime},\eta)]\right]-2||\widetilde{{Q}}_{\tau}^{(t)}-{{Q}}_{\tau}^{(t)}||_{\infty}\\
        \overset{(c)}{\ge} & {J}_{\tau}^{(t+1)}(s)-\frac{2\gamma}{1-\gamma}||\widetilde{\widehat{{Q}}}_{\tau}^{(t)}-\widehat{Q}_{\tau}^{(t)}||_{\infty}-2||\widetilde{{Q}}_{\tau}^{(t)}-{{Q}}_{\tau}^{(t)}||_{\infty}
    \end{align*}
    where $(a)$ is by applying Eq.~\eqref{eq:breve_pi_1_t+1}, $(b)$ uses the same reasoning as in Eq.~\eqref{eq:Z_2^t}, and $(c)$ is due to Eq.~\eqref{eq:hatJ-lowerbound}.
    This completes the proof.
  \endproof

\proof{\textbf{of Lemma \ref{lem:approximate-NPG-linear-system}}}
The proof mainly follows from Appendix E in \cite{cen2022fast} with some minor changes due to the new definition $\omega:=1-{\color{black}\frac{\beta\tau\gamma}{\kappa_2(1-\gamma)}}$. We present it here for self-completeness.\\
    Part (i). Bounding $||\widehat{Q}^*_{\tau}+\tau\log\widetilde{\xi}_2^{(t+1)}||_{\infty}$. From the definition \eqref{eq:approximate-xi_2t}, we have
    \begin{align*}
\left\|\widehat{Q}^*_{\tau}+\tau\log\widetilde{\xi}_2^{(t+1)}\right\|_{\infty}=&\left\|\widehat{Q}^*_{\tau}+\tau\left(\omega\log\widetilde{\xi}_2^{(t)}-(1-\omega)\frac{\widetilde{\widehat{Q}}_{\tau}^{(t)}}{\tau}\right)\right\|_{\infty}\\
        =& \left\|\omega(\widehat{Q}^*_{\tau}+\tau\log\widetilde{\xi}_2^{(t)})+(1-\omega)(\widehat{Q}^*_{\tau}-\widehat{Q}^{(t)}_{\tau})+(1-\omega)(\widehat{Q}_{\tau}^{(t)}-\widetilde{\widehat{Q}}_{\tau}^{(t)})\right\|_{\infty}\\
        \le & \omega||\widehat{Q}^*_{\tau}+\tau\log\widetilde{\xi}_2^{(t)}||_{\infty}+(1-\omega)||\widehat{Q}^*_{\tau}-\widehat{Q}^{(t)}_{\tau}||_{\infty}+(1-\omega)\delta
    \end{align*}
    Part (ii). Bounding $\max_{s,\eta,a,\eta^{\prime}}(\widehat{Q}_{\tau}^{(t+1)}(s,\eta,a,\eta^{\prime})+\tau\log\widetilde{\xi}_2^{(t+1)}(s,\eta,a,\eta^{\prime}))$. From the definition \eqref{eq:approximate-xi_2t}, we have
    \begin{align*}
        &\widehat{Q}_{\tau}^{(t+1)}(s,\eta,a,\eta^{\prime})+\tau\log\widetilde{\xi}_2^{(t+1)}(s,\eta,a,\eta^{\prime})\\
        =& \widehat{Q}^{(t+1)}_{\tau}+\tau\left(\omega\log\widetilde{\xi}_2^{(t)}-(1-\omega)\frac{\widetilde{\widehat{Q}}_{\tau}^{(t)}}{\tau}\right)\\
        =&\omega(\widehat{Q}^{(t)}_{\tau}+\tau\log\widetilde{\xi}_2^{(t)})-(1-\omega)(\widetilde{\widehat{Q}}_{\tau}^{(t)}-\widehat{Q}_{\tau}^{(t)})-(\widehat{Q}^{(t)}_{\tau}-\widehat{Q}^{(t+1)}_{\tau})\\
        \overset{(a)}{\le} & \omega(\widehat{Q}^{(t)}_{\tau}+\tau\log\widetilde{\xi}_2^{(t)}) +(1-\omega)\delta+\frac{2\delta\gamma}{1-\gamma}
    \end{align*}
    where (a) is based on $||\widetilde{\widehat{{Q}}}_{\tau}^{(t)}-\widehat{Q}_{\tau}^{(t)}||_{\infty}\le \delta$ and Eq.~\eqref{eq:approximate-Q-improvement}. Using the definition $\omega=1-{\color{black}\frac{\beta\tau\gamma}{\kappa_2(1-\gamma)}}$, we have
    \begin{align*}
        \max_{s,\eta,a,\eta^{\prime}}\left(\widehat{Q}_{\tau}^{(t+1)}(s,\eta,a,\eta^{\prime})+\tau\log\widetilde{\xi}_2^{(t+1)}(s,\eta,a,\eta^{\prime})\right)\le \omega\max(\widehat{Q}^{(t)}_{\tau}+\tau\log\widetilde{\xi}_2^{(t)}) +(1-\omega)\delta\left(1+{\color{black}\frac{2\kappa_2}{\beta\tau}}\right).
    \end{align*}
    Part (iii). Bounding $||\widehat{Q}^*_{\tau}-\widehat{Q}_{\tau}^{(t+1)}||_{\infty}$.  Because $\pi_2^{(t+1)}(\cdot,\cdot|s,\eta)=\frac{\widetilde{\xi}_2^{(t+1)}(s,\eta,\cdot,\cdot)}{||\widetilde{\xi}_2^{(t+1)}(s,\eta,\cdot,\cdot)||_1}$, we have
    \begin{align*}
        &\widehat{Q}_{\tau}^{(t+1)}(s_i,\eta_i,a_i,\eta_{i+1})-\widehat{Q}^*_{\tau}(s_i,\eta_i,a_i,\eta_{i+1})\\
        =&\gamma\mathbb{E}_{s_{i+1}\sim P(\cdot|s_i,a_i)}\left[\tau\log\left\|\exp\left(\frac{-\widehat{Q}^*_{\tau}(s_{i+1},\eta_{i+1},\cdot,\cdot)}{\tau}\right)\right\|_1-\tau\log\left\|\widetilde{\xi}_2^{(t+1)}(s_{i+1},\eta_{i+1},\cdot,\cdot)\right\|_1\right]\\
        &+\gamma\mathbb{E}_{\substack{s_{i+1}\sim P(\cdot|s_i,a_i)\\(a_{i+1},\eta_{i+2})\sim\pi_2(\cdot,\cdot|s_{i+1},\eta_{i+1})}}\left[\widehat{Q}_{\tau}^{(t+1)}(s_{i+1},\eta_{i+1},a_{i+1},\eta_{i+2})+\tau\log\widetilde{\xi}_2^{(t+1)}(s_{i+1},\eta_{i+1},a_{i+1},\eta_{i+2})\right]\\
        \overset{(a)}{\le} & \gamma||\tau\log\widetilde{\xi}_2^{(t+1)}+\widehat{Q}^*_{\tau}||_{\infty}+\gamma \max_{s,\eta,a,\eta^{\prime}}\left(\widehat{Q}_{\tau}^{(t+1)}(s,\eta,a,\eta^{\prime})+\tau\log\widetilde{\xi}_2^{(t+1)}(s,\eta,a,\eta^{\prime})\right)\\
        \overset{(b)}{\le} & \gamma\omega \max_{s,\eta,a,\eta^{\prime}}(\widehat{Q}_{\tau}^{(t)}(s,\eta,a,\eta^{\prime})+\tau\log\widetilde{\xi}_2^{(t)}(s,\eta,a,\eta^{\prime}))+\gamma (1-\omega)\delta\left(1+{\color{black}\frac{2\kappa_2}{\beta\tau}}\right)+\gamma\omega||\tau\log\widetilde{\xi}_2^{(t)}+\widehat{Q}^*_{\tau}||_{\infty}\\
        &+\gamma(1-\omega)||\widehat{Q}^*_{\tau}-\widehat{Q}_{\tau}^{(t)}||_{\infty}+(1-\omega)\gamma\delta
    \end{align*}
    where $(a)$ is due to Eq.~\eqref{eq:softmax-property1}, and $(b)$ uses Parts (i) and (ii).
    This completes the proof.
  \endproof

\proof{\textbf{of Theorem \ref{thm:linear-inexact-NPG}}}
    We start by computing the eigenvalues and eigenvectors of the matrix $B$. Specifically, the three eigenvalues of $B$ are
        $\lambda_1=\omega+\gamma(1-\omega)=1-{\color{black}\frac{\beta\tau\gamma}{\kappa_2}},\ \lambda_2=\omega,\ \lambda_3=0$
    with the corresponding eigenvectors below
    \begin{align}
        v_1=\begin{pmatrix}
            \gamma\\1\\0
        \end{pmatrix}, v_2=\begin{pmatrix}
            0\\-1\\1
        \end{pmatrix}, v_3=\begin{pmatrix}
            \omega\\ \omega-1\\0
        \end{pmatrix} \label{eq:eigenvectors}
    \end{align}
    Following Appendix E in \cite{cen2022fast}, one can show that
    \begin{align*}
        z_0\le&
        \begin{pmatrix}
            ||\widehat{Q}^*_{\tau}-\widehat{Q}_{\tau}^{(0)}||_{\infty}\\
        ||\widehat{Q}^*_{\tau}+\tau\log\widetilde{\xi}_2^{(0)}||_{\infty}\\
        ||\widehat{Q}_{\tau}^{(0)}+\tau\log\widetilde{\xi}_2^{(0)}||_{\infty}
        \end{pmatrix}\\
        = &\frac{1}{1-{\color{black}\frac{\beta\tau\gamma}{\kappa_2}}}\left((1-\omega)||\widehat{Q}^*_{\tau}-\widehat{Q}_{\tau}^{(0)}||_{\infty}+\omega\left(||\widehat{Q}^*_{\tau}+\tau\log\widetilde{\xi}_2^{(0)}||_{\infty}+||\widehat{Q}_{\tau}^{(0)}+\tau\log\widetilde{\xi}_2^{(0)}||_{\infty}\right)\right)v_1\\
        &+||\widehat{Q}_{\tau}^{(0)}+\tau\log\widetilde{\xi}_2^{(0)}||_{\infty}v_2+c_zv_3\\
        \le &\frac{1}{1-{\color{black}\frac{\beta\tau\gamma}{\kappa_2}}}\left(||\widehat{Q}^*_{\tau}-\widehat{Q}_{\tau}^{(0)}||_{\infty}+2\omega\tau||\log\pi_2^{(0)}-\log\pi_{\tau,2}^*||_{\infty}\right)v_1+||\widehat{Q}_{\tau}^{(0)}+\tau\log\widetilde{\xi}_2^{(0)}||_{\infty}v_2+c_zv_3
    \end{align*}
    where $c_z=\frac{1}{1-{\color{black}\frac{\beta\tau\gamma}{\kappa_2}}}||\widehat{Q}^*_{\tau}-\widehat{Q}_{\tau}^{(0)}||_{\infty}-\frac{\gamma}{1-{\color{black}\frac{\beta\tau\gamma}{\kappa_2}}}(||\widehat{Q}^*_{\tau}+\tau\log\widetilde{\xi}_2^{(0)}||_{\infty}+||\widehat{Q}_{\tau}^{(0)}+\tau\log\widetilde{\xi}_2^{(0)}||_{\infty})$.
    Now using the recursion \eqref{eq:approximate-recursion} and the identity $b=(1-\omega)\delta\left[(2+{\color{black}\frac{2\kappa_2}{\beta\tau}})v_1+(1+{\color{black}\frac{2\kappa_2}{\beta\tau}})v_2\right]$, we have
    \begin{align*}
        z_{t+1}\le & B^{t+1}z_0+\sum_{s=0}^t B^{t-s}b\\
        \le & B^{t+1}\left[\frac{1}{1-{\color{black}\frac{\beta\tau\gamma}{\kappa_2}}}\left(||\widehat{Q}^*_{\tau}-\widehat{Q}_{\tau}^{(0)}||_{\infty}+2\omega\tau||\log\pi_2^{(0)}-\log\pi_{\tau,2}^*||_{\infty}\right)v_1+||\widehat{Q}_{\tau}^{(0)}+\tau\log\widetilde{\xi}_2^{(0)}||_{\infty}v_2+c_zv_3\right]\\
        &+(1-\omega)\delta\sum_{s=0}^{t}B^{t-s}\left[(2+{\color{black}\frac{2\kappa_2}{\beta\tau}})v_1+(1+{\color{black}\frac{2\kappa_2}{\beta\tau}})v_2\right]\\
        =& \left[\lambda_1^t(||\widehat{Q}^*_{\tau}-\widehat{Q}_{\tau}^{(0)}||_{\infty}+2\omega\tau||\log\pi_2^{(0)}-\log\pi_{\tau,2}^*||_{\infty})+(1-\omega)\delta(2+{\color{black}\frac{2\kappa_2}{\beta\tau}})\frac{1-\lambda_1^{t+1}}{1-\lambda_1}\right]v_1\\
        &+\left[\lambda_2^{t+1}||\widehat{Q}_{\tau}^{(0)}+\tau\log\widetilde{\xi}_2^{(0)}||_{\infty}+(1-\omega)\delta(1+{\color{black}\frac{2\kappa_2}{\beta\tau}})\frac{1-\lambda_2^{t+1}}{1-\lambda_2}\right]v_2
    \end{align*}
    Since the first two entries of the eigenvector $v_2$ are non-positive, we can safely drop the terms involving $v_2$ and obtain
    \begin{align*}
        &\begin{pmatrix}
        ||\widehat{Q}^*_{\tau}-\widehat{Q}_{\tau}^{(t+1)}||_{\infty}\\
        ||\widehat{Q}^*_{\tau}+\tau\log\widetilde{\xi}_2^{(t+1)}||_{\infty}
    \end{pmatrix}\\
    \le &\left\{\lambda_1^t(||\widehat{Q}^*_{\tau}-\widehat{Q}_{\tau}^{(0)}||_{\infty}+2\omega\tau||\log\pi_2^{(0)}-\log\pi_{\tau,2}^*||_{\infty})+(1-\omega)\delta(2+{\color{black}\frac{2\kappa_2}{\beta\tau}})\frac{1-\lambda_1^{t+1}}{1-\lambda_1}\right\}\begin{pmatrix}
        \gamma\\1
    \end{pmatrix}\\
    \le & \left\{(1-{\color{black}\frac{\beta\tau\gamma}{\kappa_2}})^t(||\widehat{Q}^*_{\tau}-\widehat{Q}_{\tau}^{(0)}||_{\infty}+2\omega\tau||\log\pi_2^{(0)}-\log\pi_{\tau,2}^*||_{\infty})+\frac{2\delta}{1-\gamma}(1+{\color{black}\frac{\kappa_2}{\beta\tau}})\right\}\begin{pmatrix}
        \gamma\\1
    \end{pmatrix}\\
    =& \left\{(1-{\color{black}\frac{\beta\tau\gamma}{\kappa_2}})^tC_1+C_4\right\}\begin{pmatrix}
        \gamma\\1
    \end{pmatrix},
    \end{align*}
    where $C_1=||\widehat{Q}_{\tau}^*-\widehat{Q}_{\tau}^0||_{\infty}+2\omega\tau||\log\pi_2^{(0)}-\log\pi_{\tau,2}^*||_{\infty}$, and $C_4=\frac{2\delta}{1-\gamma}(1+{\color{black}\frac{\kappa_2}{\beta\tau}})$.
    This proves Assertion (i) in Theorem \ref{thm:linear-inexact-NPG}.
    Since $\pi_2^{(t+1)}(\cdot,\cdot|s,\eta)=\frac{\widetilde{\xi}^{(t+1)}_2(s,\eta,\cdot,\cdot)}{||\widetilde{\xi}^{(t+1)}_2(s,\eta,\cdot,\cdot)||_1}$ and $\pi^*_{\tau,2}(\cdot,\cdot|s,\eta)=\frac{\exp(-\widehat{Q}_{\tau}^*(s,\eta,\cdot,\cdot)/\tau)}{\left\|\exp(-\widehat{Q}_{\tau}^*(s,\eta,\cdot,\cdot)/\tau)\right\|_1}$, by Eq.~\eqref{eq:softmax-property}, we have
    \begin{align*}
      ||\log\pi_{\tau,2}^*-\log\pi_2^{(t+1)}||_{\infty}\le \frac{2}{\tau}||\widehat{Q}_{\tau}^*+\tau\log\widetilde{\xi}_2^{(t+1)}||_{\infty}\le \frac{2}{\tau}((1-{\color{black}\frac{\beta\tau\gamma}{\kappa_2}})^tC_1+C_4)
    \end{align*}
    This proves Assertion (ii) in Theorem \ref{thm:linear-inexact-NPG}.
    According to Eq.~\eqref{eq:Q_t1}, we have
    \begin{align}
       || Q_{\tau}^*-Q_{\tau}^{(t+1)}||_{\infty}\le \gamma(\tau||\log\pi_2^{(t+1)}-\log\pi_{\tau,2}^*||_{\infty}+||\widehat{Q}_{\tau}^{(t+1)}-\widehat{Q}_{\tau}^*||_{\infty})\le \gamma(2+\gamma)((1-{\color{black}\frac{\beta\tau\gamma}{\kappa_2}})^tC_1+C_4)\label{eq:approximate-Q-t}
    \end{align}
    This proves Assertion (iii) in Theorem \ref{thm:linear-inexact-NPG}.
    Using a similar argument as in Eq.~\eqref{eq:xi_1_recursion}, we have
    \begin{align*}
        &||{Q}_{\tau}^*+\tau\log\widetilde{\xi}_1^{(t+1)}||_{\infty}\nonumber
        \\
        \le & {\color{black}(1-\frac{\beta\tau}{\kappa_1})}||{Q}_{\tau}^*+\tau\log\widetilde{\xi}_1^{(t)}||_{\infty}+{\color{black}\frac{\beta\tau}{\kappa_1}}||{Q}_{\tau}^*-\widetilde{Q}_{\tau}^{(t)}||_{\infty}\nonumber\\
        {\le}& {\color{black}(1-\frac{\beta\tau}{\kappa_1})}^{t+1}||{Q}_{\tau}^*+\tau\log\widetilde{\xi}_1^{(0)}||_{\infty}+{\color{black}\frac{\beta\tau}{\kappa_1}}||{Q}_{\tau}^*-\widetilde{Q}_{\tau}^{(t)}||_{\infty}+{\color{black}(1-\frac{\beta\tau}{\kappa_1})}{\color{black}\frac{\beta\tau}{\kappa_1}}||{Q}_{\tau}^*-\widetilde{Q}_{\tau}^{(t-1)}||_{\infty}\nonumber\\
        &+{\color{black}(1-\frac{\beta\tau}{\kappa_1})}^2{\color{black}\frac{\beta\tau}{\kappa_1}}||{Q}_{\tau}^*-\widetilde{Q}_{\tau}^{(t-2)}||_{\infty}+\cdots+{\color{black}(1-\frac{\beta\tau}{\kappa_1})}^t{\color{black}\frac{\beta\tau}{\kappa_1}}||{Q}_{\tau}^*-\widetilde{Q}_{\tau}^{(0)}||_{\infty}\\
        \overset{(a)}{\le}& {\color{black}(1-\frac{\beta\tau}{\kappa_1})}^{t+1}||{Q}_{\tau}^*+\tau\log\widetilde{\xi}_1^{(0)}||_{\infty}+\delta+{\color{black}\frac{\beta\tau}{\kappa_1}}||{Q}_{\tau}^*-{Q}_{\tau}^{(t)}||_{\infty}+{\color{black}(1-\frac{\beta\tau}{\kappa_1})}{\color{black}\frac{\beta\tau}{\kappa_1}}||{Q}_{\tau}^*-{Q}_{\tau}^{(t-1)}||_{\infty}\nonumber\\
        &+{\color{black}(1-\frac{\beta\tau}{\kappa_1})}^2{\color{black}\frac{\beta\tau}{\kappa_1}}||{Q}_{\tau}^*-{Q}_{\tau}^{(t-2)}||_{\infty}+\cdots+{\color{black}(1-\frac{\beta\tau}{\kappa_1})}^t{\color{black}\frac{\beta\tau}{\kappa_1}}||{Q}_{\tau}^*-{Q}_{\tau}^{(0)}||_{\infty}\\
        \overset{(b)}{\le}& {\color{black}(1-\frac{\beta\tau}{\kappa_1})}^{t+1}||{Q}_{\tau}^*+\tau\log\widetilde{\xi}_1^{(0)}||_{\infty}+\delta+{\color{black}(1-\frac{\beta\tau}{\kappa_1})}^t\frac{\beta\tau}{\kappa_1}||{Q}_{\tau}^*-{Q}_{\tau}^{(0)}||_{\infty}+\frac{\gamma(2+\gamma)(1-\frac{\beta\tau\gamma}{\kappa_2})^{t}}{1-\frac{\kappa_1}{\kappa_2}\gamma}C_1+\gamma(2+\gamma)C_4\\
        \overset{(c)}{\le}& {\color{black}(1-\frac{\beta\tau\gamma}{\kappa_2})}^{t+1}||{Q}_{\tau}^*+\tau\log\widetilde{\xi}_1^{(0)}||_{\infty}+\delta+{\color{black}(1-\frac{\beta\tau\gamma}{\kappa_2})}^t\frac{\beta\tau}{\kappa_1}||{Q}_{\tau}^*-{Q}_{\tau}^{(0)}||_{\infty}+\frac{\gamma(2+\gamma)(1-\frac{\beta\tau\gamma}{\kappa_2})^{t}}{1-\frac{\kappa_1}{\kappa_2}\gamma}C_1+\gamma(2+\gamma)C_4
    \end{align*}
    where $(a)$ is true because $||{Q}_{\tau}^*-\widetilde{Q}_{\tau}^{(t)}||_{\infty}\le ||Q_{\tau}^*-Q_{\tau}^{(t)}||_{\infty}+||Q_{\tau}^{(t)}-\widetilde{Q}_{\tau}^{(t)}||_{\infty}\le ||Q_{\tau}^*-Q_{\tau}^{(t)}||_{\infty}+\delta$ and  $\delta\frac{\beta\tau}{\kappa_1}(1+(1-\frac{\beta\tau}{\kappa_1})+(1-\frac{\beta\tau}{\kappa_1})^2+\cdots+(1-\frac{\beta\tau}{\kappa_1})^t)=\delta\frac{\beta\tau}{\kappa_1}\frac{1-(1-\frac{\beta\tau}{\kappa_1})^{t+1}}{1-(1-\frac{\beta\tau}{\kappa_1})}\le \delta$, $(b)$ uses Eq.~\eqref{eq:approximate-Q-t} {\color{black} and $(c)$ uses the fact that $1-\frac{\beta\tau}{\kappa_1}<1-\frac{\beta\tau\gamma}{\kappa_2}$.}

    Finally, because $\pi^*_{\tau,1}(\cdot|s_1)\propto \exp(-{Q}_{\tau}^*(s_1,\cdot)/\tau)$ and $\pi_1^{(t+1)}(\cdot|s_1)\propto \exp(\log\widetilde{\xi}_1^{(t+1)}(s,\cdot,\cdot))$, according to Eq.~\eqref{eq:softmax-property}, we have
    \begin{align*}
       & ||\log\pi_{\tau,1}^*-\log\pi_1^{(t+1)}||_{\infty}\\
        \le& \frac{2}{\tau}||Q_{\tau}^{*}+\tau\log\widetilde{\xi}_1^{(t+1)}||_{\infty}\\
        \le & \frac{2}{\tau} \left({\color{black}(1-\frac{\beta\tau\gamma}{\kappa_2})}^{t+1}||{Q}_{\tau}^*+\tau\log\widetilde{\xi}_1^{(0)}||_{\infty}+\delta+{\color{black}(1-\frac{\beta\tau\gamma}{\kappa_2})}^t\frac{\beta\tau}{\kappa_1}||{Q}_{\tau}^*-{Q}_{\tau}^{(0)}||_{\infty}+\frac{\gamma(2+\gamma)(1-\frac{\beta\tau\gamma}{\kappa_2})^{t}}{1-\frac{\kappa_1}{\kappa_2}\gamma}C_1+\gamma(2+\gamma)C_4\right)
    \end{align*}
    Using $\omega\le 1-\beta\tau{\color{black}(1-\gamma)}$, we obtain Assertion (iv) in Theorem \ref{thm:linear-inexact-NPG}. This completes the proof.  
\endproof

\section{Detailed Algorithm}\label{append-alg}
In this appendix, we present the details of the risk-averse NPG algorithm with neural network approximation in Algorithm \ref{alg:npg}.
\allowdisplaybreaks
     \begin{algorithm}[ht!]\footnotesize
\caption{Risk-Averse NPG with Neural Network Approximation}
\begin{algorithmic}[1]  \label{alg:npg}   
\STATE Initialize neural networks for policy $\pi_1(a_1,\eta_2|s_1)$ with parameter $\theta_1$ and policy $\pi_2(a_t,\eta_{t+1}|s_t,\eta_t)$ with parameter $\theta_2$.
\WHILE{not converged}
\STATE Generate one trajectory on policy $\pi^{\theta}=(\pi_1^{\theta_1},\pi_2^{\theta_2})$: $s_1,a_1,\eta_2,c_1,s_2,\ldots,s_{T-1},a_{T-1},\eta_T,c_{T-1},s_T$.
\STATE Modify immediate costs as
$\bar{c}_1=c_1+\gamma\lambda\eta_2,\ \bar{c}_t=\frac{\lambda}{\alpha}[c_t-\eta_t]_++(1-\lambda)c_t+\gamma\lambda\eta_{t+1},\ \forall t\ge 2$.
\STATE Compute discounted costs: $V_t=\sum_{\tau=t}^{T-1}\gamma^{\tau-t}\bar{c}_{\tau}$ for all $t=1,\ldots,T-1$.
\STATE Compute the Fisher information matrix
$\mathcal{F}^{\theta_1}_{\rho}:=(\nabla_{\theta_1}\log\pi_1(a_1,\eta_2|s_1)(\nabla_{\theta_1}\log\pi_1(a_1,\eta_2|s_1))^{\mathsf T},\ \mathcal{F}^{\theta_2}_{\rho}:=\sum_{t=2}^{T-1}(\nabla_{\theta_2}\log\pi_2(a_t,\eta_{t+1}|s_t,\eta_t)(\nabla_{\theta_2}\log\pi_2(a_t,\eta_{t+1}|s_t,\eta_t))^{\mathsf T}$.
\STATE Update 
$\theta_1:=\theta_1 - \beta(\mathcal{F}^{\theta_1}_{\rho})^{-1}(\nabla_{\theta_1}\log\pi_1^{\theta_1}(a_1,\eta_2|s_1)V_1)$.
\STATE Update
$\theta_2:=\theta_2-\beta(\mathcal{F}^{\theta_2}_{\rho})^{-1}(\sum_{t=2}^{T-1}\nabla_{\theta_2}\log\pi_2^{\theta_2}(a_t,\eta_{t+1}|s_t,\eta_t)V_t)$.
\ENDWHILE
\end{algorithmic}
    \end{algorithm} 

\end{document}